\documentclass[11pt,a4paper]{article}

\usepackage{times,latexsym}
\usepackage{url}
\usepackage[T1]{fontenc}

\usepackage[acceptedWithA]{tacl2021v1}

\usepackage{nert}

\usepackage{xspace,mfirstuc,tabulary}

\newif\iftaclinstructions
\taclinstructionsfalse 
\iftaclinstructions
\renewcommand{\confidential}{}
\renewcommand{\anonsubtext}{(No author info supplied here, for consistency with
TACL-submission anonymization requirements)}
\newcommand{\instr}
\fi

\iftaclpubformat 

\else

\fi

\usepackage{makecell}
\usepackage{multirow}
\usepackage{times}
\usepackage{latexsym}
\usepackage{array}

\usepackage{adjustbox}
\usepackage{tikz}
\usetikzlibrary{arrows,shapes,trees,positioning,decorations.pathreplacing,backgrounds,arrows.meta,fit,shapes.geometric,calc,shadows}
\usepackage{tikz}
\usetikzlibrary{tikzmark,arrows.meta,calc}
\tikzset{
  movearrow/.style={-{Stealth[length=2mm]},semithick,rounded corners=1.2ex}
}
\usepackage{tikz}
\usetikzlibrary{calc}
\usepackage{booktabs}
\usepackage{caption}
\tikzset{
  wh/inline/.style={remember picture,baseline,inner sep=0pt,anchor=base},
  wh/overlay/.style={remember picture,overlay},
  wh/arrow/.style={-latex,rounded corners=.25em}
}

\newcommand{\wh}[2]{\tikz[wh/inline]\node(#1){#2};}

\newcommand{\whlink}[3][]{%
  \begin{tikzpicture}[wh/overlay]
    \draw[wh/arrow]
      (#2.south) -- +(0,-0.5em) -| (#3.south)
      node[pos=0.25,fill=white,inner sep=1pt]{\small\textit{#1}};
  \end{tikzpicture}%
}

\usepackage{adjustbox}
\usepackage{tikz-qtree}
\usetikzlibrary{arrows.meta}
\usepackage{booktabs}
\usepackage{amsmath, amssymb}
\usepackage[most]{tcolorbox}
\usepackage[nocenter]{qtree}
\usepackage[dvipsnames]{xcolor}
\usepackage[table]{xcolor}
\usepackage[T1]{fontenc}
\usepackage{forest}
\usepackage{array}
\newcolumntype{P}[1]{>{\raggedright\arraybackslash}p{#1}}
\usepackage{bbding}

\usepackage[utf8]{inputenc}

\usepackage{microtype}

\usepackage{inconsolata}

\usepackage{graphicx}
\usepackage{expex}
\usepackage{xspace}
\usepackage{colortbl,xcolor}
\usepackage[table,xcdraw]{xcolor}

\usepackage[pro]{fontawesome5}

\definecolor{poshGreen}{HTML}{089392}
\definecolor{poshYellow}{HTML}{EACF65}
\definecolor{poshRed}{HTML}{CF597E}

\usepackage[table]{xcolor}

\definecolor{ScoreLow}{HTML}{F9F9F9}
\definecolor{ScoreMid}{HTML}{E6F1FF}
\definecolor{ScoreHigh}{HTML}{AACAF5}

\newcounter{tabex} 

\renewcommand{\thetabex}{\arabic{tabex}} 

\newcommand{\tabitem}{\leavevmode\refstepcounter{tabex}(\thetabex)~}

\newcommand{\sd}[1]{\textcolor{gray}{\tiny~~(#1)}}

\newcommand{\scorecell}[1]{%
  \pgfmathparse{int(round(#1))}%
  \ifdim #1 pt < 50pt
    \cellcolor{ScoreLow}{#1}%
  \else\ifdim #1 pt < 65pt
    \cellcolor{ScoreMid}{#1}%
  \else
    \cellcolor{ScoreHigh}{#1}%
  \fi\fi
}

\newcommand{\babyfiltered}{\textbf{\textsc{\textcolor[HTML]{5898D6}{baby$_{\text{\faFilter}}$}}}\xspace} 

\newcommand{\baby}{\textbf{\textsc{\textcolor[HTML]{023FA5}{baby$_{\text{\faPlus}}$}}}\xspace} 

\newcommand{\wiki}{\textcolor[HTML]{5F5F5F}{\textsc{wiki}}\xspace}

\newcommand{\poshbench}{\textsc{PoSH-Bench}\xspace} 

\usepackage{pgf}
\usepackage{nert}

\title{A Unified Assessment of the Poverty of the Stimulus Argument\\ for Neural Language Models}

\author{
    Xiulin Yang\textsuperscript{$1$} \quad
    Arianna Bisazza\textsuperscript{$2$} \quad
    Nathan Schneider\textsuperscript{$1$} \quad
    Ethan Gotlieb Wilcox\textsuperscript{$1$}
    \\[1ex]
    \textsuperscript{$1$}Georgetown University \quad
    \textsuperscript{$2$}University of Groningen
    \\
    {
     \texttt{\{xy236, nathan.schneider, ethan.wilcox\}@georgetown.edu} \quad
        \texttt{a.bisazza@rug.nl}
    }
}

\date{}

\begin{document}
\maketitle
\begin{abstract}
Several recent contributions have evaluated the Poverty of the Stimulus Hypothesis (PoSH) using Artificial Neural Networks (ANNs). The results suggest that ANN-based language models can acquire certain structure-dependent generalizations from limited input without structural inductive biases of the type traditionally hypothesized by linguists. However, existing studies have largely focused on individual phenomena and adopted different evaluation protocols, leaving it unclear whether previous findings generalize across phenomena and learning conditions. We introduce \poshbench, a unified benchmark covering four canonical PoS phenomena. Training Transformer, LSTM, and $n$-gram models, we find that ANN-based models can achieve above-chance generalization from surprisingly limited input (10M words), but they show less efficient learning than children as input scale grows. Moreover, cognitively motivated inductive biases substantially improve broad syntactic competence but do not consistently translate to PoS-relevant generalization. Our findings provide systematic evidence challenging the claim that innate syntax is the only possible route to generalization, while suggesting that human-like learning efficiency requires inductive biases beyond those implemented here. \footnote{Models: \url{https://huggingface.co/collections/xiulinyang/posh-bench}; Code: \url{https://github.com/xiulinyang/posh-bench}}
\end{abstract}
\section{Introduction}

Despite substantial variation in their cultural and linguistic environments, children still reliably acquire their native language within a limited developmental timeframe \citep[by age 5;][]{gleitman1995invention}.
This robustness has long posed a central puzzle for theories of language acquisition and motivated the argument from
Poverty of the Stimulus (PoS) for nativist perspectives on language learning \citep{chomsky1965aspects}. The logic is straightforward: if the input is too sparse to distinguish the correct generalization rule from all other plausible alternatives, learners must possess language-specific biases that guide them toward the correct hypothesis
\citep{chomsky1976reflections}. 
PoS arguments are supported by specific linguistic phenomena---here dubbed \emph{PoS phenomena}---for which the child input is argued to be impoverished.

Such nativist claims now face an empirical challenge from neural language models. Modern transformer models acquire impressive syntactic competence from text alone, passing various grammaticality tests
\citep{warstadt-etal-2020-blimp-benchmark,wilcox2024using} without explicit linguistic constraints. However, most neural models typically train on billions of words---orders of magnitude more than a child encounters by age five (less than 100M words; \citealt{warstadt2022artificial,frank2023bridging}). Perhaps the stimulus is indeed impoverished at child scale, and neural models succeed only by compensating with massive data. This raises our central question: Can neural learners acquire PoS phenomena as reliably as children when trained on input matching the quantity and quality of children's linguistic experience?\footnote{Many human learners acquire language in visually-grounded multimodal environments. However, multimodality is not \emph{necessary} for typical language development, as blind children can acquire language without visual input \citep{landau2009language}. Due to limited cognitively aligned datasets, we focus on text-only input in this study.}

We are not alone in posing this question; indeed, many recent efforts have attempted to answer this question empirically.
A growing body of work, including recent BabyLM Challenges \citep{conll-2023-babylm, conll-2024-babylm, charpentier2025babylm}, trained language models on controlled input at human-scale data sizes and evaluated them using paradigms inspired by developmental research \citep[e.g.,][]{yedetore-etal-2023-poor, evanson-etal-2023-language}. However, three critical gaps limit our ability to use these models as valid tests of the PoS argument.

First, existing work exhibits a mismatch between developmentally plausible training conditions and systematic targeted evaluation on PoS phenomena. Human acquisition research centers on early childhood (ages 3–5), corresponding to approximately 10–50M words of exposure \citep[e.g.,][]{frank2023bridging, hart1992american}. However, computational models are typically trained on cognitively plausible data that fall outside this window, ranging from limited scales ($\le$10M words) to larger 100M-word datasets \citep[e.g.,][]{conll-2023-babylm}. This dichotomy leaves the critical 10–50M word range, which aligns most closely with the \textit{Poverty of the Stimulus} acquisition period, underexplored. Furthermore, existing benchmarks (e.g., BLiMP \citep{warstadt-etal-2020-blimp-benchmark}, Zorro \citep{huebner-etal-2021-babyberta}, SCaMP \citep{mccoy2025modeling}) assess broad linguistic generalization but do not systematically target canonical PoS phenomena. Even PoS-focused studies often examine individual phenomena in isolation, limiting cross-phenomenon comparison. Bridging this gap thus requires a unified framework that evaluates learners within the 10–50M word window against a dedicated PoS benchmark.

Second, the role of indirect evidence in acquiring PoS phenomena remains untested under developmentally plausible conditions. The PoS argument rests on the premise that (positive) direct evidence is too sparse to support generalization, necessitating innate constraints \citep{yang2016price,legate2002empirical}. Recent computational work challenges this, suggesting that indirect evidence may be richer than previously assumed \citep{misra-mahowald-2024-language,patil-etal-2024-filtered,yao2025both,foraker2009indirect}.\footnote{By \emph{direct evidence} we mean examples of a specific phenomenon. By \emph{indirect evidence} we mean examples of related phenomena, for example learning rules about topicalization based on rules about wh-question formation, where both are cases of left-dislocation in English.} However, previous studies usually assess models using adult-directed text \citep[e.g.,][]{patil-etal-2024-filtered} or do not target canonical PoS phenomena. To rigorously test whether the stimulus is truly impoverished, we must test at developmentally plausible data scales and explicitly manipulate the availability of direct positive evidence and evaluate whether models can generalize in its absence.

Third, the inductive biases that may facilitate PoS generalization remain unclear. While prior work has proposed and validated potential structural biases that improve linguistic competence overall \citep[e.g.,][]{hu-etal-2025-circuits,yoshida-etal-2024-tree}, we lack systematic evidence about (i) what sorts of biases facilitate PoS generalization, specifically, and (ii) whether biases that improve general syntactic competence also help with PoS-specific learning. This is crucial for understanding what kinds of innate constraints, if any, are needed to learn rare syntactic phenomena, or make generalizations about phenomena for which no direct positive evidence exists.

\begin{figure}[t]
    \centering
\includegraphics[width=\linewidth]{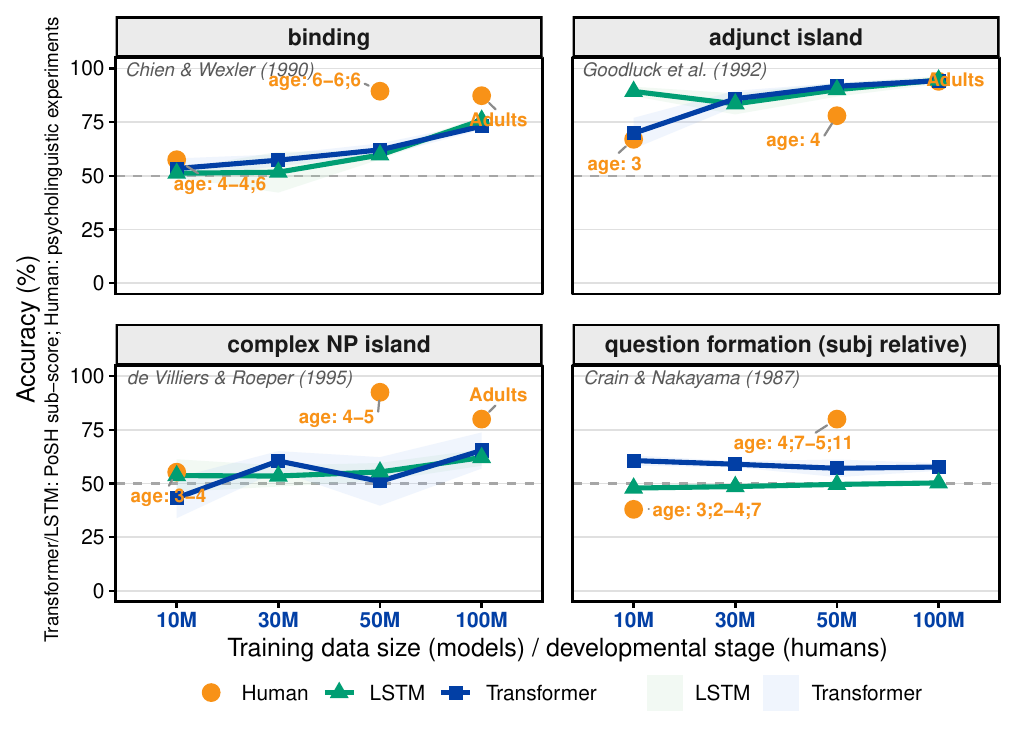}
   \caption{
Performance of human vs. neural learners. Model training sizes are aligned with the estimated cumulative input of children at specific developmental stages. Child data is from previous acquisition studies (shown in each panel); adult data is from BLiMP human judgments \citep{warstadt-etal-2020-blimp-benchmark}. Differences in evaluation protocols between behavioral studies and model evaluations preclude direct quantitative comparison; however, trajectories highlight that children acquire most phenomena with greater data efficiency.}
    \label{fig:fig1}
\end{figure}

To address these gaps, we introduce \textsc{\poshbench} (\textbf{P}overty \textbf{o}f the \textbf{S}timulus \textbf{H}ypothesis \textbf{Bench}mark), a suite designed to combine canonical PoS phenomena (islands, question formation, binding, and \textit{wanna}) and child-scale input with the explicit manipulation of direct positive evidence. We use \poshbench to answer the following three questions:
(1)~To what extent can neural or statistical models acquire PoS phenomena from developmentally plausible input?
(2)~How do input type, data scale, and direct positive evidence affect learnability?
(3)~Do certain cognitively grounded inductive biases facilitate such learning? 
In answering these questions, we do not take 100\% accuracy as an indicator of successful acquisition, since human learners also make mistakes and have individual differences. Instead, we examine (i) whether LMs show structure-dependent sensitivity to tested sentences and (ii) whether their learning trajectories develop in a manner similar to those observed in children and adults. These criteria allow us to test two crucial claims underlying the PoS argument: the first assesses whether the linguistic input provides sufficient evidence for learning, while the second assesses whether inductive biases are necessary for achieving human-like learning trajectories.

Our study provides the first systematic investigation of English syntactic PoS phenomena across multiple model classes and data scales. We find that PoS-relevant generalization is learnable from surprisingly limited input (\Cref{sec:exp1}), but human-like acquisition cannot be explained by input quantity or the amount of direct evidence alone (\Cref{fig:fig1}, \Cref{sec:exp2}). Moreover, we experiment with three types of inductive biases, two linguistic-specific and one domain-general, and we find that while these cognitive biases improve \emph{general} syntactic competence, crucially, they fail to \emph{consistently} close the efficiency gap for the phenomena targeted in \poshbench (\Cref{sec:exp3}). Instead, different biases selectively benefit different phenomena. Our findings suggest that human-like acquisition requires mechanisms beyond both distributional learning and the structural constraints explored and implemented here. 

\section{Background \& Related Work}

\subsection{The Heart of the Learnability Debate}

Children acquire complex linguistic knowledge from limited data that could, in principle, support multiple abstract grammatical hypotheses.
Yet they consistently converge on a single (correct) generalization.
For example, parents do not explicitly tell children that a noun phrase cannot be extracted from an adjunct clause (e.g., \textit{*Who did he trust Jack after seeing?}), yet children reliably acquire this constraint. Broadly speaking, two theoretical approaches, nativism and empiricism, offer competing explanations \citep{pearl2022poverty}: does syntax acquisition require innate, domain-specific constraints, or is the input itself sufficiently informative for the correct generalizations to be learned from broader domain-general mechanisms?

Nativist approaches distinguish four kinds of evidence based on two dimensions: positive vs.\ negative and direct vs.\ indirect \citep{pearl2022poverty}. 
\emph{Positive evidence} signals which forms are grammatical, while \emph{negative evidence} signals which are \emph{un}grammatical.
\emph{Direct evidence} explicitly targets the correct hypothesis; \emph{indirect evidence} requires inference from context or co-occurrence patterns. Although many linguistic phenomena can be acquired from direct evidence, others lack such evidence yet are nevertheless acquired by children. For these phenomena, the absence of direct negative evidence (e.g., explicit correction) \citep[][\emph{inter alia}]{brown1970derivational,bowerman1988no,legate2002empirical} and the sparsity or total lack of direct positive evidence \citep{legate2002empirical} forces learners to rely primarily on indirect positive input that is compatible with multiple hypotheses, giving rise to the \emph{poverty of the stimulus}.

Empiricism, in contrast, argues that the input is sufficiently rich when processed by powerful, domain-general mechanisms such as distributional learning \citep{aslin2012statistical} and pragmatic inference \citep{ambridge2011child}. Under this view, linguistic structures are learned from the data with general learning capabilities rather than from language-specific innate constraints.
\begin{figure}[!th]
    \centering
\tcbset{colback=gray!10!white, colframe=black,  
        fonttitle=\bfseries, coltitle=white, 
        boxrule=0.5pt, arc=4pt}
\begin{tcolorbox}[title=The Poverty of the Stimulus Argument]
\small
\textbf{(1) Goal:} A learner \(L\) aims to acquire the target linguistic knowledge \(T\) by identifying the correct generalization \(h_0\) from a hypothesis space
\[
H = \{h_0, h_1, \ldots, h_n\}.
\]
\textbf{(2) Constraint:} The linguistic input available to learners is finite, noisy, and sometimes misleading or ambiguous \citep{pearl2022poverty}, providing insufficient evidence to eliminate most competing hypotheses in \(H\).\\
\textbf{(3) Observation:} Despite this underdetermination, learners reliably acquire \(h_0\).  \\
\textbf{(4) The logical problem:} It is not obvious how a purely domain-general learning procedure could reliably differentiate \(h_0\) from the other hypotheses in \(H\) given the available input. \\
\textbf{(5) Conclusion:} Learners must possess domain-specific, linguistic biases that guide them systematically toward \(h_0\).
\end{tcolorbox}
\end{figure}

Following the linguistic nativist tradition, the PoS argument \citep{chomsky1965aspects} can be formally characterized as above, drawing on \citet{pearl2022poverty,pullum2002empirical,perfors2011learnability}.

\subsection{PoS phenomena Studied}
\label{sec:litreview}
In this section, we provide an overview of the four phenomena included in \poshbench.
We argue that these structures constitute the core group of linguistic phenomena that are traditionally treated as candidates for PoS-style arguments.
\footnote{We considered several additional PoS-related phenomena but excluded them for theoretical or methodological reasons. See \Cref{sec:excluded} for discussion of these phenomena.}
Specific examples can be found in \Cref{tab:pos_summary}.

\subsubsection{Yes/No Question Formation (QF)} 
\paragraph{Definition \& Human Evidence}Yes/no question formation is a classic PoS case that concerns how children learn to form yes/no questions by moving the structurally main auxiliary rather than a linearly first one \citep{chomsky1965aspects,lightfoot1991set,pullum2002empirical}, as shown in (\ref{example-qf}--\ref{example-rr}) in \Cref{tab:pos_summary}.  
To learn the correct rule, children would need sentences containing multiple auxiliaries with the main auxiliary following the subordinate auxiliary, which are extremely rare (less than 0.1\% of sentences in child-directed speech (CDS); \citealp{legate2002empirical}). 
However, children aged 4;7 or older\footnote{Following the convention in developmental studies, we use a \textsc{year; month} convention for ages.} consistently apply the structural rule \citep{crain1987structure}. 

\paragraph{Computational Modeling}
Computational studies of question formation have produced mixed results, largely depending on task formulation, learning objective, and inductive biases. Early statistical learners (e.g., $n$-grams and simple Bayesian models) exhibit surface-level success \citep{reali2005uncovering, perfors2011learnability}, but fail to generalize beyond the training distribution or across sentence types \citep{kam2008bigrams}. Neural approaches, typically framed as seq2seq mapping, also fail without explicit structural biases \citep{mccoy2018revisiting, mccoy-etal-2020-syntax, petty2021transformers, 
yedetore-etal-2023-poor}. More recent work suggests that alternative objectives and training conditions can improve performance: language modeling objectives \citep{ahuja-etal-2025-learning}, extended training regimes \citep{murty-etal-2023-grokking}, syntactically enriched data \citep{qin-etal-2025-data}, and semantic supervision \citep{yedetore-kim-2024-semantic} all yield gains. However, most positive results rely on synthetic data, limiting their cognitive plausibility. Additionally, the variation in experimental designs precludes direct comparison across studies. 
\begin{table*}[!th]
\setcounter{tabex}{0}
\small
\centering
\begin{adjustbox}{max width=\textwidth}
\begin{tabular}{l|p{2cm}|p{3cm}lp{7.5cm}}

\toprule
\textbf{Phenomenon} & \textbf{\#DPE} & \textbf{Age of Acquisition} & \textbf{Cats.} & \textbf{Minimal Pairs} \\
\midrule
\multirow{6}{*}{Question Formation} & \multirow{3}{2cm}{<0.1\% \citep{legate2002empirical}}
  &  \multirow{6}{3cm}{\raggedright 3;2--5;11 \citep{crain1987structure}; 3--5 \citep{nakayama1987performance}} & SubjR & \tabitem \label{example-qf} Will the man who \emph{did} read the book \_\_ leave? \\ 
 & & & &  \hspace{4mm}*Did the man who \_\_ read the table \emph{will} leave? \\
  \cmidrule(lr){4-5} 
&  & & ObjR & \tabitem \label{example-or} Will the man who the boy \emph{did} see \_\_ leave? \\
& \multirow{3}{2cm}{0.22\% (our study)}  & & & \hspace{4mm}*Did the man who the boy \_\_ see \emph{will} leave? \\ 
\cmidrule(lr){4-5} 
&  & & Reduced ObjR & \tabitem \label{example-rr} Can the man the boy \emph{did} see \_\_ explain? \\
 &  & & & \hspace{4mm}*Did the man the boy \_\_ see \emph{can} explain? \\  
\midrule
\multirow{6}{*}{Island Constraints} &\multirow{6}{3cm}{0}
  & \multirow{6}{3cm}{3 \citep{de1990acquisition,hirzel2022island,goodluck1992adjunct}; 3;1--6;1 \citep{de1995relative}; 4--5 \citep{fetters2017early}; 3;7--6;11 \citep{de1990acquisition}} 
 & Complex NP & \tabitem \label{example-complex-np}  \wh{wh2}{What} did Ruth who helped the girl [$_{\textsc{vp}}$ like \wh{wh1}{\_\_}]? \whlink[\CheckmarkBold]{wh1}{wh2} 
  \vspace{0.2cm}\\ 
 &  & & & \hspace{4mm}*\wh{wh2}{What} did [$_{\textsc{np}}$ Ruth who helped \wh{wh1}{\_\_}]  like the girl? \whlink[\XSolidBrush]{wh1}{wh2} 
  \vspace{0.2cm}\\
  \cmidrule(lr){4-5}
&  & & Wh & \tabitem \label{example-wh} \wh{wh2}{What} did the girl see [$_{\textsc{decl}}$ that he likes \wh{wh1}{\_\_}]? \whlink[\CheckmarkBold]{wh1}{wh2} \vspace{0.2cm}
  \\
&  & & & \hspace{4mm}*\wh{wh2}{What} did the girl see [$_{\textsc{wh}}$ whether he likes \wh{wh1}{\_\_}]? \whlink[\XSolidBrush]{wh1}{wh2} 
  \vspace{0.2cm}\\
  \cmidrule(lr){4-5}
 &  & & Adjunct & \tabitem \label{example-complex-aj} \wh{wh2}{Who} did he trust \wh{wh1}{\_\_} [$_{\textsc{adv}}$ after  seeing Jack]? \whlink[\CheckmarkBold]{wh1}{wh2} 
  \vspace{0.2cm}\\
 &  & & & \hspace{4mm}*\wh{wh2}{Who} did he trust Jack [$_{\textsc{adv}}$ after seeing \wh{wh1}{\_\_}]? \whlink[\XSolidBrush]{wh1}{wh2} \vspace{0.2cm} \\
\midrule
\multirow{4}{*}{Binding} & \multirow{3}{2cm}{0.067\% (our study)}
  & \multirow{4}{3cm}{3--5 \citep{crain1985acquisition}; 4--6;6 \citep{chien1990children}} 
  & Principle-A-L & \tabitem \label{example-complex-al} [The boy]$_i$ said [the girl]$_j$ likes herself$_{*i/j}$. \\ 
  & & & & \hspace{4mm}*[The boy]$_i$ said [the girl]$_j$ likes himself$_{i/*j}$. \\ 
  \cmidrule(lr){4-5}
  & & & Principle-A-C & \tabitem \label{example-complex-ac} [The boy]$_i$ said [[the girl]$_j$'s dad]$_k$ likes himself$_{*i/*j/k}$. \\ 
  & & & & \hspace{4mm}*[The boy]$_i$ said [[the girl]$_j$'s dad]$_k$ likes herself$_{*i/j/*k}$.\\ 
\midrule

\multirow{2}{*}{\textit{Wanna}} & 0
  & \multirow{2}{3cm}{3;11 \citep{hwang2024wanna}; 3;6 \citep{thornton1990adventures}} 
  & Wanna &\tabitem \label{example-complex-wanna}  When do you wanna laugh? \\ 
  &  & & & \hspace{5mm}*Who do you wanna laugh? \vspace{0.3cm} \\
\bottomrule
\end{tabular}
\end{adjustbox}
\caption{Phenomena included in \poshbench. \#DPE refers to the amount of direct positive evidence where available. Age of Acquisition summarizes prior studies on children’s acquisition. The benchmark covers 4 phenomena with 9 subcategories in total. We show one example for each subcategory. }
\label{tab:pos_summary}
\end{table*}
\subsubsection{Island Constraints}
\paragraph{Definition \& Human Evidence}
Island constraints \citep{ross1967constraints} are structural restrictions on dependencies between fillers (\emph{wh}-words) and gaps (empty syntactic positions). Fillers cannot associate with gaps in complex noun phrases, \mbox{\textit{wh}-clauses}, and adjunct clauses (see examples~(\ref{example-complex-np}-\ref{example-complex-aj})), among other locations. Islands constitute classic PoS cases because learners must acquire these complex constraints without access to explicit negative evidence, faced with input that would be compatible with unconstrained dependencies. However, experimental studies show that children exhibit sensitivity to certain island constraints as early as age 3 \citep{de1990acquisition,de1995relative,goodluck1992adjunct,hirzel2022island}.

\paragraph{Computational Modeling}
Island constraints appear in broad-coverage linguistic benchmarks, but performance on island-related items is often averaged with other phenomena. Targeted studies report largely positive results: pretrained language models capture various complex island effects \citep{chowdhury-zamparelli-2018-rnn, wilcox2024using, howitt-etal-2024-generalizations} but fail to show sensitivity to known exceptions to islands, such as parasitic gaps and across-the-board movement \citep{lan2024large}. Recent experiments with models trained on human-scale data \citep{chang-etal-2025-mind} further show that models trained on 10--100M words of babyLM data fail to acquire wh- and adjunct islands, though under more strict evaluation settings.

\subsubsection{Binding}
\paragraph{Definition \& Human Evidence}
Under Principle A of Binding Theory \citep{chomsky1993lectures}, an anaphor (e.g., \textit{herself}) must be locally bound by an antecedent that c-commands it as shown in (\ref{example-complex-al}--\ref{example-complex-ac}). However, disambiguating evidence, i.e., sentences with a reflexive and multiple potential antecedents, including illicit ones, are extremely rare (see \Cref{filter}). Yet children reach near-adult accuracy by around age 6 \citep{chien1990children}, with even earlier mastery reported cross-linguistically \citep{mckee1992comparison,emond2025knowledge}.

\paragraph{Computational Modeling}
Although binding is included in several syntactic benchmarks, there has been little work testing whether models can generalize Principle A from indirect evidence alone.


\subsubsection{\textit{Wanna}-Contraction}

\paragraph{Definition \& Human Evidence}
In English, \textit{want to} cannot contract to \textit{wanna} when the subject of the infinitival clause has been extracted  (\textit{Who do you want \_ to laugh?} vs.\ *\textit{Who do you wanna laugh?}), though it is permitted when extraction targets the object of \textit{want} (\textit{Who do you want to kiss \_?} $\rightarrow$ \textit{Who do you wanna kiss \_?}). Crucially, there is no direct negative evidence indicating the ungrammaticality of the former pattern. Nevertheless, children show adult-like acceptability judgments by age 3;11 \citep{hwang2024wanna}, although production errors may persist beyond comprehension \citep{getz2019acquiring,zukowski2011wanna}. To our knowledge, only one study \citep{noh-song-2026-babylms} explores this phenomenon.

\subsection{Summary} To summarize, prior work has largely studied PoS phenomena in isolation using inconsistent training and evaluation schemes. The lack of a unified framework precludes direct comparison across phenomena, leaving the validity of the PoS hypothesis for neural models an open question. 

\section{\poshbench: A Training and Evaluation Suite}

Our benchmark comprises two components: a training dataset and an evaluation suite.

\subsection{Training Data}
\label{sec:training_data}
\textsc{\poshbench} provides developmentally realistic training input that approximates both the quantity and genres of linguistic experience available to children.  
Empirical estimates of children's cumulative exposure by age 5 vary substantially across studies (15--55M in \citealp{hart1992american}; 10--35M in \citealp{gilkerson2017mapping}; and 6--60M in \citealp{frank2023bridging}).  
To balance these findings, we construct three developmentally motivated scales, i.e., 10M (lower bound), 30M (midrange), and 50M (upper bound), plus a 100M-word extension representing early adolescence \citep{macwhinney2000childes}. Prior work examines training data size and language modeling quality \citep{zhang-etal-2021-need,hu-etal-2020-systematic}, but misses the 10–50M range or relies on adult-directed data.

Children’s linguistic input is not limited to CDS: they also overhear conversations \citep{casillas2020early,cristia2019child,floor2006can,akhtar2005robustness}, television programs \citep{linebarger2005infants}, and others \citep{montag2015words}. We therefore sample a broad range of speech transcriptions rather than restricting to strictly child-directed utterances. Transcribed speech makes up roughly 76\% of the datasets from 10M to 50M. The remaining portion draws from simplified written sources (e.g., Simple English and TinyStories \citep{eldan2023tinystories}) to approximate shared reading and media exposure.\footnote{The 100M split is constructed from the BabyLM 2025 corpus \citep{charpentier2025babylm}, excluding children’s own utterances in CHILDES and replacing them with TinyStories. TinyStories consists of AI-generated short narratives designed to approximate simplified written input \citep{eldan2023tinystories}. As a result, the 100M split contains a higher proportion of written text than the 10M–50M splits, reflecting later stages of language exposure.} Compared to the BabyLM data, our training corpus has a higher proportion of speech text in the 10-50M scale (75\% vs. 58\%; \citet{charpentier2025babylm}); compared to other training corpora extracted from CHILDES \citep[e.g.,][]{yedetore-etal-2023-poor}, our training suite has a broader coverage of data size span.

\begin{table}[t]
\small
\centering
\begin{adjustbox}{width=0.48\textwidth}
\begin{tabular}{l|rrrr}
\toprule
\textbf{Data Source} & \textbf{10M} & \textbf{30M} & \textbf{50M} & \textbf{100M} \\
\midrule
\multicolumn{5}{l}{\textit{Speech Transcriptions}} \\
CHILDES   & 4M & 7M & 9M & 9M \\
OpenSubtitles & 2M & 9M & 20M & 20M \\
BNC & 1M & 5M & 8M & 8M \\
Switchboard & 0.5M & 1M & 1M & 1M \\
\rowcolor{gray!15}
\textit{Subtotal} & 7.5M & 22M & 38M & 38M \\
\midrule
\multicolumn{5}{l}{\textit{Written Texts}} \\
TinyStories & 1M & 3M & 4M & 22M \\
Project Gutenberg & 1M & 3M & 4M & 26M \\
Simple English Wikipedia & 0.5M & 2M & 4M & 14M \\
\rowcolor{gray!15}
\textit{Subtotal} & 2.5M & 8M & 12M & 62M \\
\bottomrule
\end{tabular}
\end{adjustbox}
\caption{Word counts (in millions) by data source and total target size for each training dataset.}
\label{tab:poshdata}
\end{table}

Building on this base corpus (\textsc{baby-base}; see \Cref{tab:poshdata}), we construct three training conditions. To isolate the effect of direct positive evidence while controlling for corpus size, we create two sentence-matched variants of the child-directed corpus.\footnote{Detailed statistics are provided in \Cref{stats}.} 
\begin{itemize}
    \item \babyfiltered: A filtered version of \textsc{baby-base} with all direct positive evidence sentences for the four target phenomena (\Cref{tab:pos_summary}) removed, simulating a learner exposed only to indirect evidence. Filtering details and sanity check in \Cref{filter}.
    \item \baby: Created from \babyfiltered by reinserting direct positive evidence examples at the same rate at which they naturally occur in \textsc{baby-base} ($\approx$0.2\% for question formation; $\approx$0.07\% for binding; see \Cref{filter} for how these rates were measured). The reinsertion process is necessary to ensure that \baby matches \babyfiltered in sentence count while preserving \textsc{baby-base}'s natural proportion of direct evidence. 
    \item \wiki: Wikipedia-based corpora of matched word counts, providing adult-directed baselines with denser syntax and more complex vocabulary. 
\end{itemize} 

We construct both \babyfiltered and \baby only for 10--50M scales, as the 100M condition exceeds developmentally plausible early acquisition, where the PoS argument is most relevant. For 100M, since no size-matched filtered counterpart is needed, we use the full, unfiltered \textsc{baby-base} corpus directly, which we also refer to as \baby for consistency of terminology across scales. 

\subsection{Evaluation Suite \& Metrics}
\poshbench covers four major categories discussed in \Cref{sec:litreview} (Question Formation, Binding, Islands, and \textit{Wanna}), spanning nine subcategories drawn from the acquisition literature (see examples in \Cref{tab:pos_summary}). Each subcategory includes 500 minimal pairs generated from templates. They have been manually verified for two rounds to ensure both syntactic contrast and semantic plausibility. \footnote{Revised examples include sentences that (i) contain words not included in the pre-selected word, (ii) contain unnatural collocations judged by native speakers or LLMs including \texttt{GPT-5} and \texttt{Claude Opus 4.8}, or (iii) uses verbs that can be either transitive or intransitive.} Reflexives are balanced by gender in the binding category. The templates used to create minimal pairs can be found in \Cref{benchmark-pip} and special considerations for each subcategory can be found in \Cref{sec:benchmark-design}.

We evaluate \texttt{GPT-5.4} on \poshbench to establish a reference performance and verify the overall quality of the benchmark. \texttt{GPT-5.4} achieves over 90\% accuracy on all nine subcategories, with seven subcategories exceeding 95\%.

Three of our PoSH phenomena are represented in various existing benchmarks (see details in \Cref{tab:benchmarks}). However, \textsc{\poshbench} differs from these previous studies in that it allows a systematic examination on PoS phenomena simultaneously under developmentally plausible settings.
To enable fair comparison across input conditions (\textsc{wiki} vs.\ \baby), all lexical items are sampled from the intersection of the top 3K most frequent words shared between the \textsc{wiki-100m} and \textsc{baby-100m} corpora, giving 1,950 word types (including punctuation types). 

\section{Methods: Model Training and Eval}

\subsection{Models and Training}
We train transformer and LSTM models due to their general good performance in language modeling and their different architecture. We also train $5$-gram models using Kneser-Ney smoothing with KenLM \citep{heafield2011kenlm} as baseline. For transformer models, we adopt the GPT-2 architecture \citep{radford2019language}, a standard transformer language model with an autoregressive training objective. For each dataset split, we train a separate tokenizer. This is important because although the \textsc{wiki-100m} and \textsc{baby-100m} corpora contain a comparable number of words, different vocabulary sizes yield substantially different token counts after tokenization. We therefore experiment with five common vocabulary sizes and select ${32,768}$, which minimizes compression-rate discrepancies across datasets (see \Cref{tab:ctc} in \Cref{ctc}).

We conduct pilot studies to find the best hyperparameter and architecture for both neural model classes (transformer and LSTM). In terms of transformers, we find that \texttt{gpt2-mini} achieves the lowest dev-set loss on 10M-scale data, whereas \texttt{gpt2-small} performs better on larger datasets. Accordingly, we use \texttt{gpt2-mini} for 10M experiments and \texttt{gpt2-small} for 30M, 50M, and 100M. Models are trained for up to 100k steps with early stopping (patience = 6k) and a 4k-step linear warmup \citep{padovani2025child}. As for LSTMs, we experiment with number of layers and embedding dimensions as well as other hyperparameters, including batch sizes, learning rate, dropout rate, warmup steps, to find the best hyperparameter setting for each split.

For each neural model class, we train each model with three random seeds. Models with the same random seed share the same tokenizer. We also evaluate model perplexity on the dev set to compare with the perplexity of the 5-gram model as a sanity check. The perplexity results can be found in Table~\ref{tab:ppl}. Pilot studies and model training hyperparameters are provided in \Cref{appendix:training}.

\subsection{Evaluation}
For each minimal pair in \poshbench, we compute sentence-level perplexity for both sentences and count an item as correct if the grammatical sentence receives lower perplexity. Accuracy (the proportion of correct preferences), averaged across three seeds, serves as the primary metric. Our implementation builds on \texttt{minicons} \citep{misra2022minicons}.

\begin{table*}[tbp]
\centering
\begin{adjustbox}{max width=1\textwidth}
\begin{tabular}{ll*{12}{l}}
\toprule
 &  & \multicolumn{3}{c}{\textbf{10M}} & \multicolumn{3}{c}{\textbf{30M}} & \multicolumn{3}{c}{\textbf{50M}} & \multicolumn{2}{c}{\textbf{100M}} &   \\
\cmidrule(lr){3-5}\cmidrule(lr){6-8}\cmidrule(lr){9-11}\cmidrule(lr){12-13}
\textbf{Model} & \textbf{Category} & \babyfiltered & \baby & \textsc{wiki} & \babyfiltered & \baby & \textsc{wiki} & \babyfiltered & \baby & \textsc{wiki} & \baby & \textsc{wiki} & \textbf{GPT2} \\
\midrule
\multirow{5}{*}{Transformer}
 & Island & \scorecell{68.2}{***}\sd{4.2} & \scorecell{67.9}{***}\sd{0.6} & \scorecell{68.0}{***}\sd{2.6} & \scorecell{72.0}{***}\sd{4.3} & \scorecell{78.9}{***}\sd{1.0} & \scorecell{76.1}{***}\sd{4.4} & \scorecell{72.2}{***}\sd{0.9} & \scorecell{76.2}{***}\sd{4.3} & \scorecell{75.7}{***}\sd{1.3} & \scorecell{81.5}{***}\sd{4.1} & \scorecell{77.5}{***}\sd{2.7} & \scorecell{95.1}{***} \\
 & Question Formation & \scorecell{55.7}{***}\sd{3.5} & \scorecell{55.0}{***}\sd{2.6} & \scorecell{61.0}{***}\sd{2.0} & \scorecell{54.7}{***}\sd{2.1} & \scorecell{53.3}{***}\sd{0.9} & \scorecell{59.2}{***}\sd{2.6} & \scorecell{53.8}{***}\sd{0.9} & \scorecell{54.1}{***}\sd{0.7} & \scorecell{58.7}{***}\sd{1.3} & \scorecell{51.9}{}\sd{0.9} & \scorecell{60.4}{***}\sd{1.5} & \scorecell{78.7}{***} \\
 & Binding & \scorecell{52.8}{*}\sd{0.9} & \scorecell{53.4}{}\sd{4.7} & \scorecell{45.7}{***}\sd{1.8} & \scorecell{57.6}{***}\sd{1.1} & \scorecell{57.3}{***}\sd{2.7} & \scorecell{63.1}{***}\sd{3.0} & \scorecell{65.1}{***}\sd{2.0} & \scorecell{62.1}{***}\sd{3.0} & \scorecell{61.7}{***}\sd{2.6} & \scorecell{73.1}{***}\sd{1.7} & \scorecell{60.7}{***}\sd{6.0} & \scorecell{77.5}{***} \\
 & Wanna & \scorecell{77.7}{***}\sd{2.4} & \scorecell{64.9}{***}\sd{11.4} & \scorecell{25.6}{***}\sd{16.0} & \scorecell{70.1}{***}\sd{13.8} & \scorecell{61.5}{***}\sd{7.2} & \scorecell{52.4}{}\sd{16.5} & \scorecell{84.9}{***}\sd{2.2} & \scorecell{75.2}{***}\sd{5.3} & \scorecell{51.4}{}\sd{4.1} & \scorecell{92.1}{***}\sd{3.4} & \scorecell{65.7}{}\sd{23.4} & \scorecell{43.6}{**} \\
\cmidrule(lr){2-14}
 & Average & \scorecell{63.6}{}\sd{11.6} & \scorecell{60.3}{}\sd{7.2} & \scorecell{50.1}{}\sd{18.8} & \scorecell{63.6}{}\sd{8.7} & \scorecell{62.8}{}\sd{11.3} & \scorecell{62.7}{}\sd{10.0} & \scorecell{69.0}{}\sd{13.0} & \scorecell{66.9}{}\sd{10.7} & \scorecell{61.9}{}\sd{10.2} & \scorecell{74.7}{}\sd{17.1} & \scorecell{66.1}{}\sd{8.0} & \scorecell{73.7}{}\sd{21.6} \\
\midrule
\multirow{5}{*}{LSTM}
 & Island & \scorecell{74.7}{***}\sd{3.1} & \scorecell{77.6}{***}\sd{2.8} & \scorecell{66.5}{***}\sd{4.0} & \scorecell{79.0}{***}\sd{3.5} & \scorecell{75.4}{***}\sd{3.2} & \scorecell{70.7}{***}\sd{1.0} & \scorecell{82.9}{***}\sd{5.9} & \scorecell{78.5}{***}\sd{0.5} & \scorecell{72.2}{***}\sd{2.3} & \scorecell{82.8}{***}\sd{1.9} & \scorecell{80.7}{***}\sd{4.3} \\
 & Question Formation & \scorecell{53.4}{***}\sd{0.9} & \scorecell{53.5}{***}\sd{1.9} & \scorecell{56.2}{***}\sd{0.8} & \scorecell{50.6}{}\sd{1.7} & \scorecell{50.3}{}\sd{0.8} & \scorecell{55.5}{***}\sd{1.3} & \scorecell{45.8}{***}\sd{0.3} & \scorecell{48.2}{*}\sd{1.6} & \scorecell{53.9}{***}\sd{3.1} & \scorecell{45.1}{***}\sd{0.7} & \scorecell{53.8}{***}\sd{2.6} \\
 & Binding & \scorecell{49.8}{}\sd{2.5} & \scorecell{51.2}{}\sd{1.9} & \scorecell{56.4}{***}\sd{0.8} & \scorecell{51.7}{}\sd{2.9} & \scorecell{51.7}{}\sd{9.5} & \scorecell{56.2}{***}\sd{1.6} & \scorecell{55.4}{***}\sd{2.5} & \scorecell{59.7}{***}\sd{2.0} & \scorecell{60.5}{***}\sd{5.1} & \scorecell{75.9}{***}\sd{1.7} & \scorecell{61.0}{***}\sd{1.0} \\
 & Wanna & \scorecell{65.3}{}\sd{19.3} & \scorecell{74.3}{***}\sd{6.0} & \scorecell{7.6}{***}\sd{3.7} & \scorecell{78.1}{***}\sd{5.6} & \scorecell{82.1}{***}\sd{8.3} & \scorecell{38.9}{***}\sd{6.4} & \scorecell{63.6}{***}\sd{6.7} & \scorecell{71.7}{***}\sd{17.2} & \scorecell{52.0}{}\sd{20.4} & \scorecell{82.1}{***}\sd{3.0} & \scorecell{53.2}{*}\sd{2.0} \\
\cmidrule(lr){2-14}
 & Average & \scorecell{60.8}{}\sd{11.4} & \scorecell{64.1}{}\sd{13.7} & \scorecell{46.7}{}\sd{26.5} & \scorecell{64.8}{}\sd{15.8} & \scorecell{64.9}{}\sd{16.3} & \scorecell{55.3}{}\sd{13.0} & \scorecell{61.9}{}\sd{15.8} & \scorecell{64.5}{}\sd{13.4} & \scorecell{59.7}{}\sd{9.1} & \scorecell{71.5}{}\sd{17.8} & \scorecell{62.2}{}\sd{12.8} & -- \\
 \midrule 
\multirow{5}{*}{N-gram} & Island & \scorecell{58.3}{***} & \scorecell{55.8}{***} & \scorecell{63.6}{***} & \scorecell{60.8}{***} & \scorecell{62.7}{***} & \scorecell{65.3}{***} & \scorecell{61.4}{***} & \scorecell{72.0}{***} & \scorecell{66.9}{***} & \scorecell{63.9}{***} & \scorecell{55.7}{***} \\
&  Question Formation & \scorecell{60.7}{***} & \scorecell{51.5}{} & \scorecell{57.5}{***} & \scorecell{56.8}{***} & \scorecell{52.7}{*} & \scorecell{58.3}{***} & \scorecell{56.9}{***} & \scorecell{50.1}{} & \scorecell{57.7}{***} & \scorecell{57.5}{***} & \scorecell{47.0}{*} \\
&  Binding & \scorecell{45.3}{**} & \scorecell{45.4}{**} & \scorecell{43.8}{***} & \scorecell{43.1}{***} & \scorecell{46.3}{*} & \scorecell{43.8}{***} & \scorecell{43.1}{***} & \scorecell{46.2}{*} & \scorecell{44.2}{***} & \scorecell{43.4}{***} & \scorecell{48.9}{} \\
&  Wanna & \scorecell{77.4}{***} & \scorecell{66.8}{***} & \scorecell{53.0}{} & \scorecell{79.6}{***} & \scorecell{58.6}{***} & \scorecell{49.8}{} & \scorecell{74.6}{***} & \scorecell{5.8}{***} & \scorecell{49.8}{} & \scorecell{73.6}{***} & \scorecell{56.2}{*} \\
\cmidrule(lr){2-14}
 & Average & \scorecell{60.4}{}\sd{13.2} & \scorecell{54.9}{}\sd{9.0} & \scorecell{54.5}{}\sd{8.3} & \scorecell{60.1}{}\sd{15.1} & \scorecell{55.1}{}\sd{7.1} & \scorecell{54.3}{}\sd{9.5} & \scorecell{59.0}{}\sd{13.0} & \scorecell{43.5}{}\sd{27.6} & \scorecell{54.6}{}\sd{9.9} & \scorecell{59.6}{}\sd{12.7} & \scorecell{52.0}{}\sd{4.7}& -- \\
\bottomrule
\end{tabular}
\end{adjustbox}
\caption{Category-level PoSH results grouped by model (rows) and training size (columns). Each category's accuracy pools all of its subcategories' pairs within a seed (e.g. Binding: 2 subcategories $\times$ 500 pairs = 1{,}000 pairs). For each seed, significance against chance (50\%) is assessed using a $\chi^2$ goodness-of-fit test. The per-seed $p$-values are then combined using Fisher's method, and the resulting $p$-values are corrected using the Holm-Bonferroni procedure. Significance is reported only when all three seeds show the same direction relative to chance. ({*}~$p < 0.05$, {**}~$p < 0.01$, {***}~$p < 0.001$).}
\label{tab:result}
\end{table*}

\section{Experiment 1: Generalization without Positive Evidence}
\label{sec:exp1}
This experiment tests whether neural and statistical models can acquire PoS phenomena from limited input in the absence of direct positive evidence.

\subsection{Experiment Setup}
We trained models at multiple data scales on \babyfiltered, using the optimal model size for each scale, and evaluated performance on \poshbench. We also evaluated GPT-2 \citep{radford2019language}, which shares the same architecture as \texttt{gpt2-small} but is trained on 40GB of data, providing an example of the same model architecture given substantially more data.

\subsection{Results}
Category-level results are shown in \Cref{tab:result} (\babyfiltered column), with subcategory-level results in \Cref{sec:finegrained} for transformers (\Cref{tab:finegrained}), LSTMs (\Cref{tab:finegrained-lstm}), and 5-gram models (\Cref{tab:finegrained-ngram}).

We find that transformer models trained on \babyfiltered are able to achieve above-chance accuracy even with only 10M words of training data. Two-sided $\chi^2$ tests at chance ($p = 0.5$) confirm that all main categories trained on \babyfiltered are significantly above chance (\Cref{tab:result}). This indicates that transformer models can extract hierarchical regularities from indirect evidence alone, which aligns with previous studies on rare phenomena \citep[e.g.,][]{misra-mahowald-2024-language, yao2025both}, and other phenomena, including binding and islands covered in BLiMP \citep{patil-etal-2024-filtered}. By contrast, LSTMs and 5-gram models cannot generalize to all phenomena. LSTMs demonstrate consistent above-chance performance only at 50M or greater data scales. And $n$-gram models show significantly \emph{below} chance behavior (e.g., for Binding tests) as well as low overall accuracy, even for 100M data scales.

It is also important to note that, although transformer models are able to generalize globally, they still struggle with specific subcategories. For example, among island constraints, complex-NP islands are harder to learn for transformer models than adjunct and wh-islands. Within question formation, the (reduced) object relative clause 
subcategories remain harder to learn than subject relative clauses, with accuracy still close to chance even at 100M words. This echoes the broader pattern documented in child relative clause processing, where object relatives are acquired later and comprehended less reliably than subject relatives \citep[e.g.,][]{friedmann2009relativized,mckee1998relatives,fluck1978comprehension}.

\section{Experiment 2: Effects of Input Type, Data Scale, and Evidence Availability}
\label{sec:exp2}

This experiment investigates how input quantity, domain, and the availability of direct positive evidence jointly affect structural generalization. We extend our training to include models trained on \wiki and \baby and conduct a statistical analysis to address three questions:
(1) Does increasing pretraining scale consistently improve generalization for these datasets?
(2) Does exposure to complex, adult-oriented text (\wiki) yield better syntactic abstraction than developmentally plausible data?
(3) Does the presence of sparse direct positive evidence ($<\!0.3\%$) facilitate generalization?

\subsection{Experiment Setup} To ensure robustness, we evaluated models not only on \poshbench but also on overlapping phenomena from existing benchmarks, specifically the overlapping Binding and Island constraints in BLiMP \citep{warstadt-etal-2020-blimp-benchmark}, Zorro \citep{huebner-etal-2021-babyberta}, and SCaMP \citep{mccoy2025modeling} (see \Cref{tab:category-dataset-map} for subcategory selection). This cross-benchmark evaluation serves as a consistency check, reducing the risk that observed effects are artifacts of a single dataset design.

We further analyzed results on \poshbench using a Bayesian binomial mixed-effects model with weakly informative priors for two neural model classes separately: \texttt{correct} $\sim$ (\texttt{size} + \texttt{category}) $*$ \texttt{source\_filtered} + (1\,|\,\texttt{seed}) + (1\,|\,\texttt{phenomenon}).\footnote{Sampling used four chains of 2{,}000 iterations each (1{,}000 warmup), yielding 4{,}000 posterior draws. Because \textsc{wiki} has no filtered counterpart, we combine \texttt{filter} and \texttt{training\_source} into a single three-level factor (\texttt{baby\_unfiltered}, \texttt{baby\_filtered}, \texttt{wiki}) rather than treating them as separate two-level predictors, which would otherwise conflate \texttt{wiki} with the unfiltered baseline and confound the filtering effect with the domain effect.} The dataset consists of 148.5k observations, where each data point represents whether the model assigned higher probabiltiy to the grammatical version (\texttt{correct}$=1$ or $0$) of each minimal pair item. All predictors are sum-coded. \looseness=-1
\subsection{Results} 

\paragraph{Overall results}  
Results are shown in the \baby and \wiki columns of \Cref{{tab:result}}; cross-dataset comparisons are visualized in \Cref{fig:overall} (\Cref{sec:finegrained}).
We find that the learning trajectories for island constraints and binding align across benchmarks, though absolute difficulty levels vary. For island constraints, \poshbench yields higher accuracy than the other three benchmarks, while for binding \poshbench is comparatively harder, likely due to template and vocabulary choices.     Learning trajectories are consistent across benchmarks.
Accuracy and rate-of-increase with data scale vary between model types. Transformers show a higher rate of increase in generalization with data scale (i.e., slope in \cref{fig:overall}). LSTMs require more complex data (\wiki 50M) to achieve generalization across all phenomena, while $5$-gram models fail to generalize even with 100M data.\looseness=-1

We also observe that all models trained on \wiki perform particularly poorly on the \textit{wanna} subcategory. This likely happens because \wiki consists primarily of written text, in which occurrences of \textit{wanna} are extremely rare (only 0.011\% of sentences in the \wiki 100M corpus contain \textit{wanna}).
 Consequently, the models are unlikely to receive sufficient evidence to learn the relevant syntactic constraint and may instead rely on lexical associations between wh-words and verbs when assigning probabilities. The same explanation may account for the below-chance performance of the pretrained GPT-2 model, which was also pretrained primarily on written text.

\paragraph{Bayesian mixed model results}

Posterior medians and 95\% credible intervals (CrI) are visualized in \Cref{fig:bayesian}, with detailed estimates reported in \Cref{tab:brmsresults}.\footnote{Convergence diagnostics for both mixed models indicate good mixing across chains ($\hat R = 1.00$ for all parameters), large effective sample sizes, and no divergent transitions.} Below, we address each of our research questions in turn:

\begin{figure}[t]
    \centering

    \begin{subfigure}{\linewidth}
        \centering
        \includegraphics[width=\linewidth]{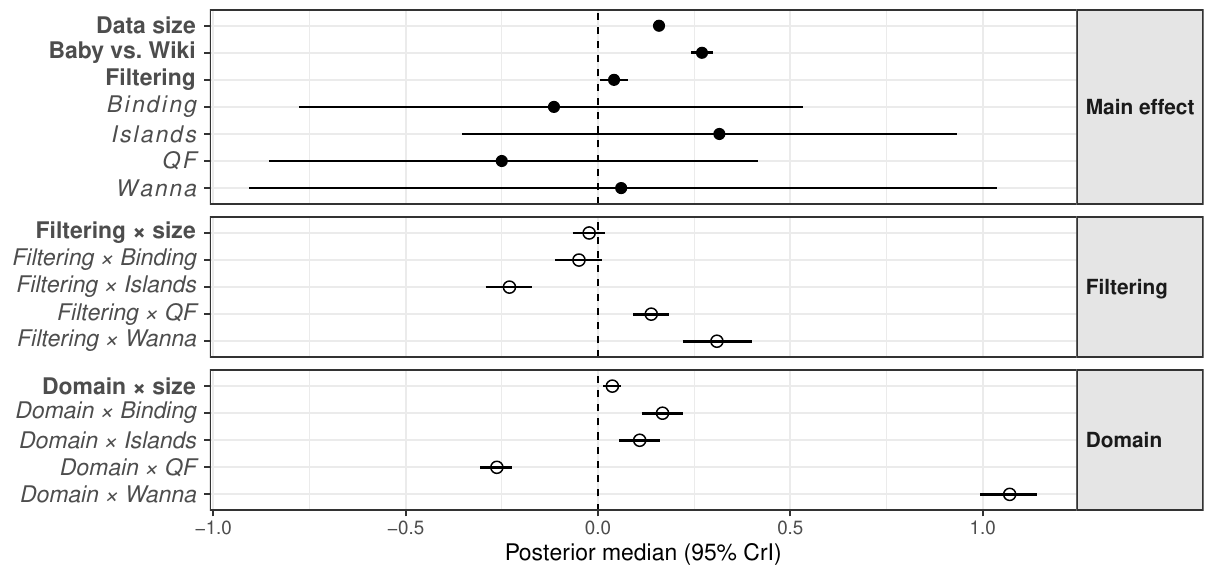}
        \caption{Transformer}
    \end{subfigure}

    \vspace{0.5em}

    \begin{subfigure}{\linewidth}
        \centering
        \includegraphics[width=\linewidth]{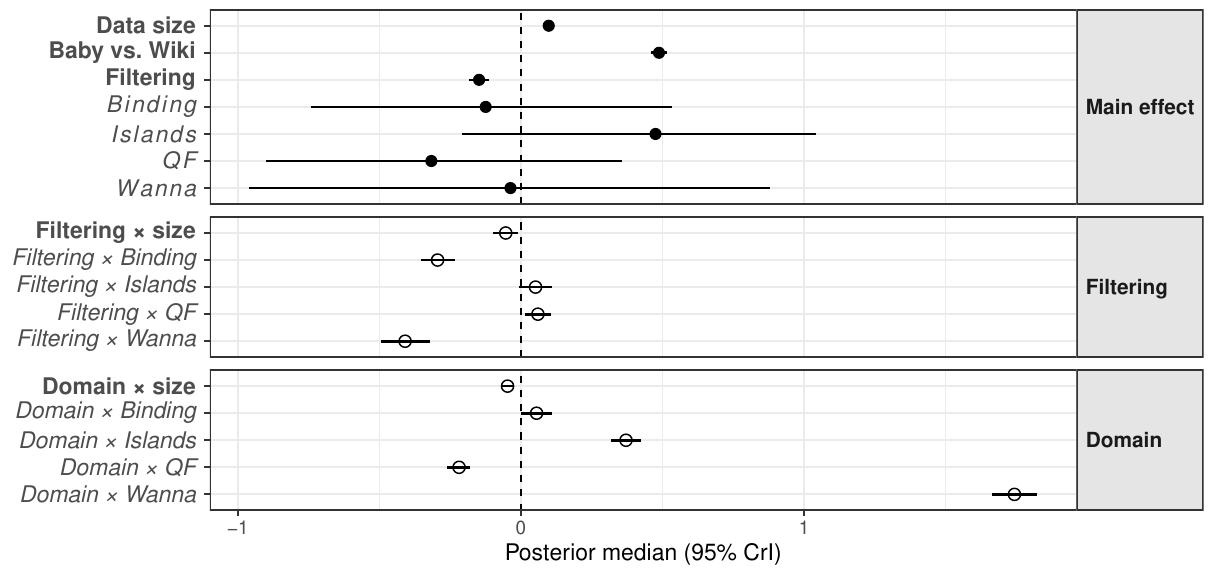}
        \caption{LSTM}
    \end{subfigure}

    \caption{Bayesian modeling results showing the effects of filtering, training corpus, data size, and PoS categories. Positive effects indicate higher scores. QF=Question Formation.}
    \label{fig:bayesian}
\end{figure}

\newcommand{\cri}[3]{(\ensuremath{#1~\textcolor{gray}{ [ #2,\,#3 ]})}\xspace}

First, data scale matters. We observe a robust positive effect of training data size for both transformer \cri{0.16}{0.14}{0.17} and LSTM models \cri{0.10}{0.08}{0.11}, confirming that larger corpora improve generalization across phenomena.

Second, developmentally plausible data proves more effective for PoS generalization. Models trained on \textsc{baby} consistently outperform those trained on \wiki for both types of neural models in binding, question formation, and \textit{wanna}. This pattern contrasts with findings by \citet{padovani2025child}, likely due to methodological differences: (i) our data sources are not limited to CDS from CHILDES and the observed benefits may arise from increased genre or syntactic diversity in child-directed corpora 
\citep{feng-etal-2024-child,qin-etal-2025-data}; (ii) we target different linguistic phenomena and use different benchmark vocabulary sampling strategies: \citet{padovani2025child} focus on subject-verb agreement and thus require fine-grained dependency relations, while  we rely on the overlap between the most frequent word lists from the two domains. Future work is needed to disentangle these confounding variables.

Third, the effect of direct positive evidence is limited and phenomenon-/model-dependent. We did not observe a consistent effect of direct positive evidence for both LSTM and transformer models. We find that Question Formation and \textit{wanna} contraction appears to benefit from the filtering of direct evidence for Question Formation and Binding, while for LSTM, \textit{wanna} shows worse results \cri{-0.41}{-0.49}{-0.32}. This suggests that although different model types are able to generalize across PoS phenomena, they might rely on different sources of information.

As for the various interaction effects observed between filtering and categories, we provide two possible explanations and leave further investigation for future work. First, unlike island constraints or question formation, which are structural patterns realized across a broad range of lexical items, \textit{wanna}-contraction is anchored to a single lexical form. Because the filtered dataset removes a vast number of standard questions, the relative proportion of the remaining questions that contain \textit{wanna} significantly increases.  Thus, for rare phenomena like \textit{wanna}, the relative prominence of evidence in the training distribution may be more critical than the sheer volume of related structural cues. Second, as for the improvement of Question Formation under filtering condition, we believe it is relevant to the definition we made regarding direct positive evidence, which we adopt from previous language acquisition studies \citep{legate2002empirical,pearl2022poverty}. It refers to sentences that support the structure-dependent hypothesis (i.e., the correct sentences in the minimal pairs) and provide evidence against the alternative, structure-independent hypothesis (i.e., the incorrect sentences in the minimal pairs). See \cref{filter} for details. As the direct positive evidence for Question Formation are more complex sentences, removing them may therefore reduce the overall syntactic complexity of the training distribution, potentially easing acquisition of the target generalization together with other indirect evidence, even as it removes disambiguating evidence.

This definition of direct positive evidence may implicitly shape the distribution of evidence available to learners during training. We cannot rule out the possibility that other sentences provide important cues for learning, even though we classify them as indirect evidence under our definition. Understanding which forms of indirect evidence are most informative for structure learning remains an open and interesting question.

\section{Experiment 3: Inductive Biases -- Helpful or Not?}
\label{sec:exp3}
We investigate whether incorporating cognitively motivated inductive biases into the models facilitates PoS generalization under limited input. Specifically, we test three biases proposed by previous work, which differ in strength and locus: (i) weak hierarchical bias via pretraining on formal languages, (ii) strong structural bias via syntactic language models, and (iii) a dynamic recency bias inspired by working memory development.

\subsection{Three Cognitively Informed Biases}
\paragraph{Weak hierarchical bias via Dyck pretraining.}

Drawing on syntactic bootstrapping theories \citep{fisher2010syntactic}, we hypothesize that prior exposure to recursive structures facilitates subsequent language acquisition. Various studies have confirmed that an initial phase of pretraining on formal languages (sometimes called ``pre-pretraining'') helps the model to converge better \citep{papadimitriou-jurafsky-2023-injecting}. Following \citet{hu-etal-2025-circuits}, we implemented the pretraining approach with a shuffled $k$-Dyck language. We pretrained the model on this formal language for 2k steps before continuing pretraining on the target \babyfiltered dataset. This manipulation instills a weak bias towards hierarchical processing without injecting explicit linguistic rules. \looseness=-1

\begin{figure*}[t]
    \centering
\includegraphics[width=0.96\linewidth]{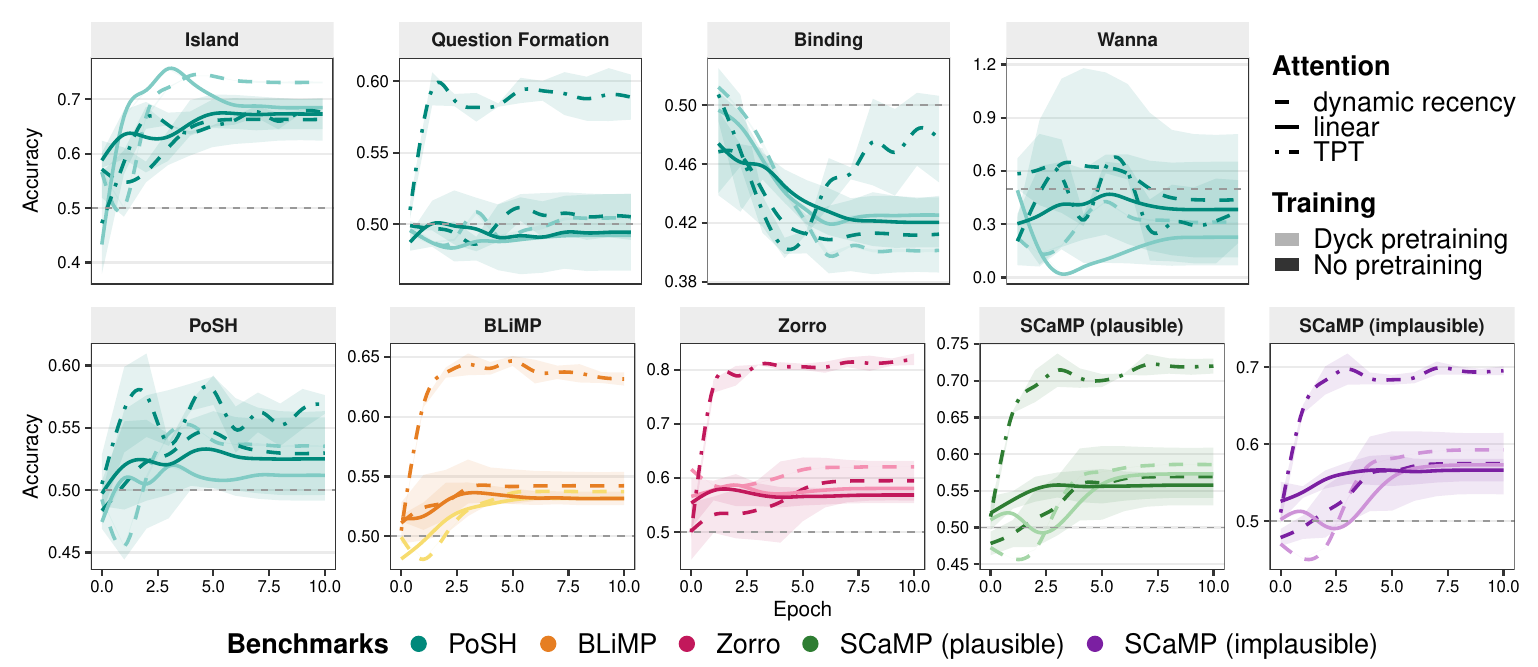}
  \caption{Results of GPT-2 mini with/without different inductive biases. Top: Per-category breakdown. Bottom: Benchmark overall averages. Shaded: $\pm1$ SD across 3 random seeds. Solid darker line: Vanilla transformer baseline without any modifications.}
\label{fig:cogexp}
\end{figure*}
\paragraph{Strong Structural Bias via Syntactic Language Models}
\citet{mccoy-etal-2020-syntax} showed that tree-based models show better hierarchical generalization. To test the effect of explicit structural guidance, we adopt a class of syntactic language models called Tree-Planted Transformer \citep[TPT; ][]{yoshida-etal-2024-tree}. Unlike standard LMs that attend to tokens solely based on content and position, TPT incorporates a tree-distance matrix into the attention mechanism. This implicitly supervises the model to align its attention distribution with syntactic phrase structure, simulating a learner who has access to (or innately infers) strict structural relations. Following \citet{yoshida-etal-2024-tree}, we used dependency parses during training, which have been shown to yield the strongest syntactic gains; all input was parsed using the spaCy parser \citep{honnibal2020spacy}.

\paragraph{Dynamic recency bias from limited working memory.}
The \textit{Less-is-More} hypothesis \citep{newport1990maturational} posits that children’s limited working memory during early development paradoxically aids generalization by filtering out complex noise and focusing attention on local constituents. To simulate this constraint, we adopt the algorithm from \citet{mita-etal-2025-developmentally}, implementing a \textit{Dynamic Recency Bias}. Specifically, the attention score for the $i$-th query is defined as $\mathrm{softmax} (\mathbf{q}_i K^\top + \textcolor{blue}{s_{decay}})$. $\mathbf{q}_i K^\top$ denotes the standard attention logits, and $\textcolor{blue}{s_{decay}}=r^t \cdot \mathbf{b}_i$ which imposes a dynamic locality penalty. $\mathbf{b}_i = [ -(i-1), \dots, 0 ]$ represents the relative distance vector specific to position $i$, and the decay rate $r$ (where $0 < r < 1$) controls the bias strength: as the epoch $t$ increases, $r^t$ decreases, gradually relaxing the constraint and expanding the effective context window.

\subsection{Experiments \& Evaluation}

We follow the same training procedure as previous studies, use GPT2-mini, and train on the \babyfiltered-10M dataset. For TPT, we train the model on sentence level, while for Dyck pretraining and Dynamic Recency, we use chunk-level training. For each experiment, we train models on 10 epochs. This fixed duration is particularly critical for the Dynamic Recency model, where the decay rate is calibrated ($r=0.6$) to allow the bias to vanish naturally by the final epoch. Regarding hyperparameters, we maintain the same optimization settings as in Experiment 1 (\Cref{tab:hyperparameter}), with the exception that no warmup steps were used. Each experimental condition was replicated with three random seeds.

Apart from testing three cognitively inspired biases alone, we also test the interaction between pretraining and dynamic recency bias and see if combining both leads to better performance.

We evaluated structural generalization on \poshbench alongside three external benchmarks: BLiMP, Zorro, and SCaMP (including both plausible and implausible conditions). This comprehensive suite allows us to: (i) verify our findings on specific PoS phenomena (e.g., islands, binding) across different test set constructions, and (ii) assess whether biases that help with PoS phenomena also generalize to broader linguistic competencies.

\subsection{Results}
Results are visualized in \Cref{fig:cogexp}. The top row plots the learning trajectories for each phenomenon in \poshbench, while the bottom row displays the macro-average accuracy for each benchmark.

We observe a clear dissociation between performance on general-purpose linguistic benchmarks and PoS-specific evaluation. On broad benchmark suites such as BLiMP, Zorro, and SCaMP, TPT consistently achieves the strongest performance, followed by models incorporating Dynamic Recency. In contrast, on \poshbench, differences among inductive biases are much smaller and less consistent, though TPT still shows a relative edge, driven primarily by gains on question formation. Although TPT has previously been reported to improve long-distance dependency processing when evaluated on SyntaxGym \citep{gauthier-etal-2020-syntaxgym}, this advantage does not extend to island phenomena. This suggests that biases optimized for broad coverage do not straightforwardly transfer to the specific data-sparse constraints of PoS phenomena.

We also find that different inductive biases favor distinct phenomena (\Cref{fig:cogexp}, top). TPT excels at binding and question formation; the binding result is confirmed across all benchmarks (Appendix \Cref{fig:islandandbinding}), while Dyck pretraining combined with dynamic recency specifically aids island constraints, a long-distance dependency phenomenon.

\section{Discussion \& Conlcusion}
\paragraph{Does the Poverty of the Stimulus Hypothesis hold for neural learners?}
As \citet[p.~1]{samet2008historical} noted, ``although [innateness] is an easy doctrine to attack, it is a hard one to kill.'' Our findings suggest a nuanced answer to the PoS hypothesis as applied to neural learners. On one hand, transformer models achieve nontrivial learning of PoS phenomena with approximately 10M words even without direct positive evidence, challenging the assumption that the input is too impoverished to support generalization without innate constraints. On the other hand, the persistent gap between model and human learning efficiency suggests that either (i) the input contains richer structure than current models can exploit, or (ii) additional factors beyond the statistical learning employed by our models facilitate human-like generalization.

This gap is visualized in \Cref{fig:fig1}, with performance on psycholinguistic tests and references to related studies given in \Cref{tab:relatedwork}.
The above trend is particularly evident in binding: while models match 4-year-olds in exceeding chance performance at 10M words \citep{chien1990children}, they plateau at $\approx$75\% even with 100M words, whereas children rapidly reach 90\% agreement in psycholinguistic tests by age 6.

\paragraph{Why do cognitively informed biases improve general competence but not PoS generalization?} We consider two possible explanations for Experiment 3. 
First, successful generalization may require coordinated deployment of multiple biases rather than any single bias in isolation. This aligns with \citeposs{mccoy-etal-2020-syntax} finding that generalization emerges from complex interactions among factors. We observe some evidence for this: Dyck pretraining plus dynamic locality bias aid the acquisition of island constraints, while TPT aids question formation and binding. This raises the question: do human children also rely on phenomenon-specific biases, or do they possess a more unified structural learning mechanism? 
 
Additionally, the same inductive bias can be realized through diverse architectural mechanisms and experimental configurations (e.g., varying training scales or modalities). In this study, we restricted our scope to specific implementations and a fixed set of hyperparameters. It is plausible that different hyperparameters or alternative implementations of structural biases, such as those in Transformer Grammars \citep{sartran-etal-2022-transformer} and PaLM \citep{peng-etal-2019-palm}, might yield different generalization patterns. However, a systematic grid search over these alternatives is computationally expensive. 

Either way, our results suggest that improvements in broad syntactic competence are relatively robust across the inductive biases we tested, whereas their effects on PoSH-relevant generalization may be more sensitive to the specific architecture, hyperparameter configuration, or inductive bias interaction.

\begin{table}[t]
\small
    \centering
   \begin{adjustbox}{max width=0.49\textwidth}
    \begin{tabular}{l|ll}
    \toprule
    Category     &  Acc (Younger) & Acc (Older) \\
    \midrule 
    adjunct island & 67\% (3) & 78\% (4)\\
    complex-NP island &  55.3\% (3--4) & 92.5\% (4--5)\\
    question formation     & 38\% (3;2--4;7) & 80\% (4;7--5;11) \\
    binding & 57.5\% (4--4;6) & 89.4\% (6--6;6) \\
    \bottomrule
    \end{tabular}
    \end{adjustbox}
    \caption{Reported accuracy for four classic syntactic phenomena at earlier vs.\ later stages of child language development between 3 and 6. 
Values are drawn from prior developmental studies including adjunct island \citep{goodluck1992adjunct}, complex-NP island \citep{de1995relative}, question formation \citep{crain1987structure}, and binding \citep{chien1990children} based on generation or act-out comprehension tasks. Other phenomena studied in our paper do not have age-related results in literature.}
\label{tab:relatedwork}
\end{table}

\paragraph{Conclusion \& Future Work}
In this work, we provide the first systematic investigation of the Poverty of the Stimulus argument through the lens of modern neural language models and a developmentally grounded benchmark, \textsc{\poshbench}. Across three experiments, we find that transformers trained on child-scale data partially acquire core linguistic generalizations but remain less data-efficient than human learners; that additional exposure to direct evidence or adult-oriented text does not substantially improve generalization; and that cognitively motivated inductive biases enhance overall linguistic competence but fail to transfer to PoS generalization.

The PoS argument is not specific to English but has been investigated across languages, e.g., island constraints in Chinese \citep{huang1998logical}, Turkish \citep{ozsoy1996dependencies}, and Japanese \citep{watanabe1992subjacency}, although the relevant constraints may vary across languages. We focus on English due to comparability with prior work and limited developmentally plausible resources in other languages. Future work will extend \textsc{\poshbench} to multilingual and multimodal settings, integrate psycholinguistic developmental data, and explore additional cognitively plausible hypotheses. There has been ongoing work exploring relevant directions \citep[e.g.,][]{jumelet-etal-2026-babybabellm,wang2025babyvlmv2developmentallygroundedpretraining,kobzeva-etal-2023-neural} but no relevant PoS benchmark exists yet. Such work will bring us closer to understanding not only how much data matters, but what kind of learner makes that data meaningful.\looseness=-1

\section*{Acknowledgments}
We thank Najoung Kim, Kanishka Misra, and members of TinLab,  PICoL, and NERT for their helpful discussion and feedback. Compute resources provided through the Department of Linguistics and the Massive Data Institute at Georgetown University were used in part of this work. Arianna Bisazza is supported by the Talent Programme of the Dutch Research Council (grant VI.Vidi.221C.009).

\bibliography{anthology}
\bibliographystyle{acl_natbib}
\newpage
\appendix
\section{Other PoS Phenomena Considered}
\label{sec:excluded}
We considered but ultimately excluded the following PoS-related phenomena:

\begin{itemize}
  \item \textbf{Anaphoric \textit{one}.} Although anaphoric \textit{one} has been widely studied \citep[e.g.,][]{floor2006can, pearl2011far, foraker2009indirect}, recent empirical work suggests that it may not be strictly constrained by the syntactic representation of its antecedent \citep{payne2013anaphoric}, making it an unstable test case for PoS arguments.\footnote{In a pilot study, we attempted to replicate native speaker judgments for anaphoric \textit{one} but observed inconsistency, aligning with the findings of \citet{payne2013anaphoric}. Given this inconsistency in adult judgments, it is unclear whether anaphoric \textit{one} constitutes a viable test case for PoS arguments.}

  \item \textbf{Control and Raising.} Control and raising constructions are typically acquired after age 5 \citep{chomsky1969acquisition, choe2016children}, falling outside the 3–5 year developmental window targeted in this study.

  \item \textbf{Binding Principle C.} Although young children show early sensitivity to Binding Principle C \citep{crain1985acquisition}, testing this phenomenon requires cross-sentential reference interpretation, introducing discourse-level confounds that are difficult to control.
\end{itemize}

More broadly, PoS arguments have also been advanced beyond syntax, including in phonology \citep{wilson2006learning}, lexical learning \citep{braine1990can, scott2009two}, morphology \citep{gordon1985level}, and semantics \citep{crain2000acquisition, falmagne2013language, papafragou2003scalar}. We limit our focus on syntactic constructions as they stand at the core of PoS.

\section{Statistics of the \baby and \textsc{wiki} datasets}
\label{stats}
The detailed statistics for our \textsc{wiki} \babyfiltered, and \baby datasets can be found in \Cref{tab:datastats}.

\section{Filtering the Corpus}
\label{filter}

This section details the procedure used to remove sentences containing \textit{direct positive evidence} for the target PoS phenomena.  
Among the five main constructions examined, two (Question Formation and Binding) have identifiable direct positive evidence that could directly support the correct generalization.  
Our goal is to construct a conservative filtering procedure that minimizes the risk of leaving potential direct positive evidence in the training corpus, even at the cost of including some false positives.

\paragraph{Parsing and General Strategy}
All sentences in the \baby corpus were parsed using Stanza \citep{qi2020stanza}.  
For the filtered condition (\babyfiltered), we applied rule-based heuristics over dependency parses and POS tags.  
To ensure coverage, rules were deliberately designed to err on the side of over-filtering (i.e., allowing more false positives rather than false negatives).

\paragraph{Question Formation}
For question formation, according to \citet{legate2002empirical,sampson1989language}, the direct positive evidence can be two types of sentences: (i) \textit{Is the boy [who is in the corner]$_{NP}$  smiling?} and (ii) \textit{How could [anyone that was awake]$_{NP}$ not hear that?} In other words, the sentences must contain a relative clause modifying the subject. In this case, the children would know that the fronted auxiliary is not the first auxiliary linearly but the first auxiliary hierarchically.

Sentences were removed if they (i) contained a relative clause modifying the subject, and (ii) were interrogative.  
For (i), we identified sentences containing both \texttt{acl:relcl} and \texttt{nsubj} dependencies where the token labeled \texttt{nsubj} precedes that labeled \texttt{acl:relcl}.  
For (ii), we detected questions based on the presence of a final question mark, a wh-word, or an auxiliary verb among the first two tokens.
We filtered out 0.22\% sentences from the Baby-base 100M corpus. 

\paragraph{Binding}
To remove direct positive evidence for reflexive binding, we excluded sentences containing a reflexive ending in \textit{self} or \textit{selves} that was preceded by at least two noun or pronoun tokens.  
POS tags for identifying reflexives and potential antecedents were extracted from Stanza parses. We filtered out 0.067\% sentences from the Baby-base 100M corpus. 

\paragraph{Sanity Check}
To validate that our filtering strategy effectively removed direct positive evidence, we performed a manual inspection on a random sample of the filtered dataset (\babyfiltered). We randomly sampled 1,000 sentences containing potential triggers for each phenomenon and examined them individually.

For \textbf{Question Formation}, we retrieved sentences containing a question mark and at least two auxiliaries (11k candidates out of 1.5M). We manually checked for subject-auxiliary inversion where the inverted auxiliary is not the linearly first auxiliary. We found zero examples meeting this condition, confirming that our filtering was successful.

For \textbf{Binding}, we retrieved sentences containing reflexives (5.7k candidates). Potential direct positive evidence consists of sentences where the correct antecedent is distinguishable from distractors only via syntactic structure (c-command) and locality constraints, rather than linear locality. 
In our 1k sample, we found only 6 examples constituting such ``leakage.'' In the list below, we bold the correct antecedents and mark potential distractors in gray to illustrate why these serve as disambiguating evidence:

\begin{itemize}
    \item \textbf{Evidence against Linear Adjacency:}
    In these cases, the correct antecedent is structural, while the linearly closest noun acts as a distractor (often due to feature mismatch or structural depth).
    \begin{itemize}
        \item \textit{\textbf{People} like \textbf{\textcolor{gray}{her}} don't often get themselves murdered.} \\
        (Distractor `her' is linearly closest but does not c-command.)
        \item \textit{As the burgher ceased, \textbf{his fellow townsmen} wept aloud, and \textbf{\textcolor{gray}{many, amid tears}}, threw themselves at his feet...} \\
        (Distractor `tears' is linearly closest but is an object of a preposition.)
        \item \textit{All we know is, that \textbf{master and \textcolor{gray}{slave}} trusted themselves alone to a small bark...} \\
        (The model must bind to the coordinate phrase rather than the closest singular noun `slave'.)
    \end{itemize}

    \item \textbf{Evidence for Locality (Binding Domain):}
    In these relative clause structures, the matrix subject c-commands the reflexive but is ruled out by the \textit{locality constraint} (Principle A). The model receives evidence that c-command alone is insufficient without domain restrictions.
    \begin{itemize}
        \item \textit{\textbf{\textcolor{gray}{Mrs. Cochran}} swayed against \textbf{David}, who pulled himself together.} \\
        (Distractor `Mrs. Cochran' c-commands the reflexive but is outside the local CP; the model must strictly bind within the relative clause.)
        \item \textit{\textbf{\textcolor{gray}{I}}'ll tell the entire story to \textbf{those folks} who think themselves so fine.} \\
        (Distractor `I' c-commands but is non-local; correct binding is to the local head `folks'.)
        \item \textit{\textbf{\textcolor{gray}{The boys}} were not allowed... except \textbf{three or four} who ... thought themselves richly paid...} \\
        (Distractor `The boys' is the matrix subject, ruled out by locality.)
    \end{itemize}
\end{itemize}

We take a leakage rate of 30  sentences in 1.5M (0.002\%) to be negligible. Additionally, this minimal signal is further compromised by the presence of exempt anaphors and logophors (e.g., reflexives in locative PPs) identified in our sample. For example:
\begin{itemize}
    \item \textit{I think maybe when people own things and then \textbf{they} pass away, a part of themselves gets printed on those things...} (Reflexive in a logophoric context; no c-commanding antecedent.)
    \item \textit{\textbf{Most animals}, instead of building a nest in front of themselves, build it round themselves.} (Reflexive in a locative PP adjunct; the antecedent is not c-commanding the anaphor and thus exempt from strict binding.)
\end{itemize}
These instances act as confounding noise, providing valid linguistic counter-evidence to strict c-command constraints. Consequently, they counteract the leaked positive evidence, ensuring that the leakage remains too sparse and inconsistent to support robust structural generalization.

\section{Corpus Token Count}
\label{ctc}
The corpus token count with different vocabluary sizes can be found in \Cref{tab:ctc}.
\section{Hyperparameter setting}
\label{appendix:training}
We experiment with models with 4 different architectures, which we call GPT2-small, GPT2-mini, and GPT2-xs and GPT2-xxs. We list their architecture information in \Cref{tab:modelconfig}. 

\begin{table}[!th]
\small 
    \centering
    \begin{tabular}{l|c}
    \toprule
     Hyperparameter    & Setting \\
    \midrule 
     learning rate    &  1e-4\\
     batch size & 32\\
     context length & 512\\
     warmup steps & 4,000\\
     steps & 100,000 \\
     patience & 6,000 \\
     dropout & 0.1\\
     weight decay & 0.1 \\
     learning rate scheduler & linear \\
     
     \bottomrule
    \end{tabular}
    \caption{Hyperparameter settings for the experiments}
    \label{tab:hyperparameter}
\end{table}

\begin{table}[th]
\small
    \centering
    \begin{adjustbox}{width=0.5\textwidth}
    \begin{tabular}{l|p{4.5cm}}
    \toprule
    Phenomenon    & Existing Benchmarks \\
    \midrule
     Question Formation    & \citet{mccoy-etal-2020-syntax,yedetore-etal-2023-poor}\\
     Island Constraints & \citet{warstadt-etal-2020-blimp-benchmark,huebner-etal-2021-babyberta,wilcox-etal-2018-rnn,mccoy2025modeling}\\
     Binding & \citet{warstadt-etal-2020-blimp-benchmark, huebner-etal-2021-babyberta,mccoy2025modeling} \\
     \textit{Wanna} & None \\
     \bottomrule
    \end{tabular}
    \end{adjustbox}
    \caption{Existing benchmarks that contain the phenomena of interest}
    \label{tab:benchmarks}
\end{table}

\begin{table}[t]
\small
    \centering
    \begin{tabular}{c|c}
    \toprule
    Category & Accuracy \\
    \midrule
    question-formation-rr     &  96.8\\
    question-formation-sr & 99.8  \\  
    question-formation-or & 99.8 \\  
    \textit{wanna} & 91.6\\
    principle-a-locality & 100\\
    principle-a-command & 99.4\\
    island-adjunct & 96.4\\
    island-wh & 93.6\\
    island-complex-np & 96.4\\
         \bottomrule
    \end{tabular}
    \caption{Results of \texttt{GPT5.4}}
    \label{tab:gpt5}
\end{table}

\begin{table}[t]
\centering
\small

\textbf{Transformer Models}

\vspace{0.3em}

\begin{adjustbox}{max width=\columnwidth}
\begin{tabular}{lcccc}
\toprule
\textbf{Parameter} & \textsc{mini} & \textsc{xs} & \textsc{xxs} & \textsc{small} \\
\midrule
Hidden size & 512 & 512 & 512 & 768 \\
\#Heads & 8 & 8 & 4 & 12 \\
\#Layers & 4 & 6 & 6 & 12 \\
FFN dim & 2048 & 2048 & 2048 & 3072 \\
\#Parameters & 29.6M & -- & -- & 110.6M \\
\bottomrule
\end{tabular}
\end{adjustbox}

\vspace{1em}

\textbf{LSTM Models}

\vspace{0.3em}

\begin{adjustbox}{max width=\columnwidth}
\begin{tabular}{lccccc}
\toprule
\textbf{Parameter} & \textsc{tiny} & \textsc{xs} & \textsc{small} & \textsc{medium} & \textsc{large} \\
\midrule
Hidden size & 640 & 768 & 1024 & 2048 & 2048 \\
\#Layers & 2 & 2 & 2 & 2 & 3 \\
\#Parameters & 27.6M & 34.6M & 50.4M & 134.3M & 167.9M \\
\bottomrule
\end{tabular}
\end{adjustbox}

\caption{Model configurations used in our experiments.}
\label{tab:modelconfig}

\end{table}

\begin{table*}[t]
\small
\centering
\begin{tabular}{l|cc|cc|cc|cc}
\toprule
Dataset 
& \multicolumn{2}{c|}{10M}
& \multicolumn{2}{c|}{30M}
& \multicolumn{2}{c|}{50M}
& \multicolumn{2}{c}{100M} \\
\cmidrule(lr){2-3}
\cmidrule(lr){4-5}
\cmidrule(lr){6-7}
\cmidrule(lr){8-9}
& \#Sents & \#Words/Sent
& \#Sents & \#Words/Sent
& \#Sents & \#Words/Sent
& \#Sents & \#Words/Sent \\
\midrule
\wiki 
& 430k & 23.2
& 1.3M & 23.52
& 2.1M & 23.41
& 4.3M & 23.51 \\
\baby
& 1.4M & 7.13
& 4.0M & 7.53
& 6.6M & 7.46
& 11.6M & 8.55 \\
\babyfiltered 
& 1.4M & 7.11
& 4.0M & 7.50
& 6.6M & 7.43
& NA & NA \\
\bottomrule
\end{tabular}
\caption{Statistics of training corpora by data size. Each size group reports the number of sentences and average sentence length.}
\label{tab:datastats}
\end{table*}

\begin{table*}[tbp]
\centering
\begin{adjustbox}{max width=1\textwidth}
\begin{tabular}{ll*{12}{l}}
\toprule
  &   & \multicolumn{3}{c}{\textbf{10M}} & \multicolumn{3}{c}{\textbf{30M}} & \multicolumn{3}{c}{\textbf{50M}} & \multicolumn{2}{c}{\textbf{100M}} &   \\
\cmidrule(lr){3-5}\cmidrule(lr){6-8}\cmidrule(lr){9-11}\cmidrule(lr){12-13}
\textbf{Category} & \textbf{Phenomenon} & \babyfiltered & \baby & \textsc{wiki} & \babyfiltered & \baby & \textsc{wiki} & \babyfiltered & \baby & \textsc{wiki} & \baby & \textsc{wiki} & \textbf{GPT2} \\
\midrule
\multirow{3}{*}{Island} & island\_adjunct & \scorecell{73.5}{***}\sd{4.5} & \scorecell{69.8}{***}\sd{7.2} & \scorecell{71.3}{***}\sd{5.7} & \scorecell{85.6}{***}\sd{5.9} & \scorecell{85.9}{***}\sd{3.7} & \scorecell{76.2}{***}\sd{7.7} & \scorecell{91.3}{***}\sd{3.3} & \scorecell{91.7}{***}\sd{1.5} & \scorecell{81.9}{***}\sd{8.7} & \scorecell{94.3}{***}\sd{1.4} & \scorecell{89.4}{***}\sd{2.2} & \scorecell{97.4}{***} \\
  & island\_complex\_np & \scorecell{37.4}{}\sd{11.8} & \scorecell{43.3}{}\sd{9.4} & \scorecell{49.1}{}\sd{0.8} & \scorecell{43.7}{}\sd{10.5} & \scorecell{60.5}{***}\sd{4.6} & \scorecell{63.5}{***}\sd{5.8} & \scorecell{41.6}{***}\sd{2.8} & \scorecell{51.0}{}\sd{11.3} & \scorecell{60.5}{***}\sd{8.8} & \scorecell{65.4}{***}\sd{8.4} & \scorecell{56.9}{***}\sd{4.1} & \scorecell{94.8}{***} \\
  & island\_wh & \scorecell{93.7}{***}\sd{2.9} & \scorecell{90.7}{***}\sd{3.4} & \scorecell{83.7}{***}\sd{6.7} & \scorecell{86.7}{***}\sd{2.0} & \scorecell{90.4}{***}\sd{5.6} & \scorecell{88.6}{***}\sd{4.8} & \scorecell{83.7}{***}\sd{3.4} & \scorecell{86.0}{***}\sd{2.8} & \scorecell{84.9}{***}\sd{1.6} & \scorecell{84.9}{***}\sd{6.0} & \scorecell{86.3}{***}\sd{4.0} & \scorecell{93.2}{***} \\
\midrule
\multirow{3}{*}{Question Formation} & question\_formation\_or & \scorecell{54.0}{}\sd{6.6} & \scorecell{54.9}{**}\sd{2.5} & \scorecell{61.9}{***}\sd{4.0} & \scorecell{53.5}{*}\sd{3.1} & \scorecell{51.7}{}\sd{2.2} & \scorecell{57.8}{***}\sd{3.2} & \scorecell{52.8}{}\sd{3.6} & \scorecell{52.8}{}\sd{0.5} & \scorecell{54.3}{*}\sd{1.1} & \scorecell{47.9}{}\sd{0.7} & \scorecell{62.1}{***}\sd{0.8} & \scorecell{81.0}{***} \\
  & question\_formation\_rr & \scorecell{51.5}{}\sd{5.1} & \scorecell{49.5}{}\sd{3.3} & \scorecell{63.3}{***}\sd{2.5} & \scorecell{50.5}{}\sd{4.4} & \scorecell{49.1}{}\sd{3.6} & \scorecell{58.7}{***}\sd{2.8} & \scorecell{49.7}{}\sd{1.5} & \scorecell{52.5}{}\sd{4.2} & \scorecell{58.3}{***}\sd{3.5} & \scorecell{50.0}{}\sd{2.5} & \scorecell{57.6}{***}\sd{2.6} & \scorecell{75.4}{***} \\
  & question\_formation\_sr & \scorecell{61.5}{***}\sd{1.3} & \scorecell{60.7}{***}\sd{2.4} & \scorecell{57.8}{***}\sd{0.5} & \scorecell{60.1}{***}\sd{0.8} & \scorecell{59.0}{***}\sd{1.4} & \scorecell{61.2}{***}\sd{3.5} & \scorecell{58.9}{***}\sd{4.0} & \scorecell{57.1}{***}\sd{4.1} & \scorecell{63.4}{***}\sd{1.2} & \scorecell{57.7}{***}\sd{1.8} & \scorecell{61.5}{***}\sd{3.9} & \scorecell{79.6}{***} \\
\midrule
\multirow{2}{*}{Binding} & principle\_a\_command & \scorecell{44.6}{**}\sd{1.4} & \scorecell{47.4}{}\sd{4.8} & \scorecell{29.9}{***}\sd{3.0} & \scorecell{43.7}{***}\sd{1.7} & \scorecell{39.7}{***}\sd{1.7} & \scorecell{45.5}{**}\sd{3.3} & \scorecell{50.3}{}\sd{3.5} & \scorecell{41.2}{***}\sd{0.5} & \scorecell{41.1}{***}\sd{4.2} & \scorecell{55.1}{***}\sd{4.0} & \scorecell{37.1}{***}\sd{9.4} & \scorecell{59.4}{***} \\
  & principle\_a\_locality & \scorecell{61.0}{***}\sd{1.4} & \scorecell{59.4}{***}\sd{4.6} & \scorecell{61.5}{***}\sd{2.2} & \scorecell{71.5}{***}\sd{3.1} & \scorecell{74.9}{***}\sd{4.2} & \scorecell{80.7}{***}\sd{2.9} & \scorecell{79.9}{***}\sd{1.3} & \scorecell{83.1}{***}\sd{5.7} & \scorecell{82.3}{***}\sd{1.0} & \scorecell{91.1}{***}\sd{1.3} & \scorecell{84.3}{***}\sd{2.5} & \scorecell{95.6}{***} \\
\midrule
\multirow{1}{*}{Wanna} & wanna & \scorecell{77.7}{***}\sd{2.4} & \scorecell{64.9}{***}\sd{11.4} & \scorecell{25.6}{***}\sd{16.0} & \scorecell{70.1}{***}\sd{13.8} & \scorecell{61.5}{***}\sd{7.2} & \scorecell{52.4}{}\sd{16.5} & \scorecell{84.9}{***}\sd{2.2} & \scorecell{75.2}{***}\sd{5.3} & \scorecell{51.4}{}\sd{4.1} & \scorecell{92.1}{***}\sd{3.4} & \scorecell{65.7}{}\sd{23.4} & \scorecell{43.6}{**} \\
\midrule
Average &   & \scorecell{61.7}{}\sd{0.2} & \scorecell{60.1}{}\sd{1.5} & \scorecell{56.0}{}\sd{1.6} & \scorecell{62.8}{}\sd{2.5} & \scorecell{63.6}{}\sd{1.8} & \scorecell{64.9}{}\sd{3.1} & \scorecell{65.9}{}\sd{0.3} & \scorecell{65.6}{}\sd{2.6} & \scorecell{64.2}{}\sd{1.2} & \scorecell{70.9}{}\sd{1.1} & \scorecell{66.8}{}\sd{2.6} & \scorecell{80.0}{} \\
\bottomrule
\end{tabular}
\end{adjustbox}
\caption{Fine-grained PoSH results grouped by training size (columns) and phenomenon (rows). The last row reports the average across all PoSH phenomena (mean and standard deviation over seeds). For each cell, a $\chi^2$ goodness-of-fit test against chance (50\%) is run per seed and the resulting $p$-values are combined via Fisher's method; the combined $p$-values are then corrected for multiple comparisons using the Holm-Bonferroni procedure within the same dataset, applied separately within each training-data condition (column), and stars are assigned based on the resulting adjusted $p$-values. ({*}~$p < 0.05$, {**}~$p < 0.01$, {***}~$p < 0.001$, significance is reported only when all three seeds show the same direction relative to chance.).}
\label{tab:finegrained}
\end{table*}

\begin{table*}[!th]
\centering
\begin{adjustbox}{max width=1\textwidth}
\begin{tabular}{ll*{11}{l}}
\toprule
  &   & \multicolumn{3}{c}{\textbf{10M}} & \multicolumn{3}{c}{\textbf{30M}} & \multicolumn{3}{c}{\textbf{50M}} & \multicolumn{2}{c}{\textbf{100M}} \\
\cmidrule(lr){3-5}\cmidrule(lr){6-8}\cmidrule(lr){9-11}\cmidrule(lr){12-13}
\textbf{Category} & \textbf{Phenomenon} & \babyfiltered & \baby & \textsc{wiki} & \babyfiltered & \baby & \textsc{wiki} & \babyfiltered & \baby & \textsc{wiki} & \baby & \textsc{wiki} \\
\midrule
\multirow{3}{*}{Island} & island\_adjunct & \scorecell{84.4}{***}\sd{1.6} & \scorecell{89.3}{***}\sd{2.2} & \scorecell{64.7}{***}\sd{13.5} & \scorecell{88.7}{***}\sd{2.0} & \scorecell{83.5}{***}\sd{4.9} & \scorecell{73.2}{***}\sd{6.0} & \scorecell{88.9}{***}\sd{0.6} & \scorecell{90.1}{***}\sd{3.2} & \scorecell{75.3}{***}\sd{5.3} & \scorecell{94.6}{***}\sd{1.8} & \scorecell{87.8}{***}\sd{5.9} \\
  & island\_complex\_np & \scorecell{46.9}{}\sd{11.2} & \scorecell{53.7}{}\sd{7.6} & \scorecell{54.7}{**}\sd{3.1} & \scorecell{56.5}{}\sd{9.6} & \scorecell{53.5}{}\sd{4.9} & \scorecell{56.3}{}\sd{6.6} & \scorecell{65.5}{***}\sd{14.9} & \scorecell{55.4}{***}\sd{4.2} & \scorecell{60.9}{}\sd{10.8} & \scorecell{62.1}{***}\sd{3.8} & \scorecell{70.5}{***}\sd{3.7} \\
  & island\_wh & \scorecell{92.9}{***}\sd{1.5} & \scorecell{89.7}{***}\sd{1.2} & \scorecell{80.3}{***}\sd{2.4} & \scorecell{91.8}{***}\sd{3.0} & \scorecell{89.3}{***}\sd{5.0} & \scorecell{82.7}{***}\sd{1.5} & \scorecell{94.3}{***}\sd{3.1} & \scorecell{90.0}{***}\sd{4.3} & \scorecell{80.3}{***}\sd{8.6} & \scorecell{91.7}{***}\sd{1.9} & \scorecell{83.7}{***}\sd{3.8} \\
\midrule
\multirow{3}{*}{Question Formation} & question\_formation\_or & \scorecell{53.5}{}\sd{1.4} & \scorecell{58.0}{***}\sd{1.6} & \scorecell{57.9}{***}\sd{1.8} & \scorecell{55.8}{***}\sd{1.2} & \scorecell{52.4}{}\sd{0.5} & \scorecell{54.5}{**}\sd{3.8} & \scorecell{46.5}{*}\sd{1.2} & \scorecell{48.4}{}\sd{2.5} & \scorecell{55.5}{***}\sd{2.8} & \scorecell{41.5}{***}\sd{0.6} & \scorecell{53.3}{}\sd{4.2} \\
  & question\_formation\_rr & \scorecell{53.2}{}\sd{0.2} & \scorecell{54.6}{**}\sd{3.7} & \scorecell{60.1}{***}\sd{1.5} & \scorecell{48.0}{}\sd{1.9} & \scorecell{50.0}{}\sd{2.9} & \scorecell{59.7}{***}\sd{1.4} & \scorecell{45.4}{**}\sd{0.6} & \scorecell{46.5}{}\sd{1.5} & \scorecell{55.1}{***}\sd{4.3} & \scorecell{43.6}{***}\sd{2.4} & \scorecell{53.8}{}\sd{1.6} \\
  & question\_formation\_sr & \scorecell{53.4}{}\sd{1.6} & \scorecell{47.9}{}\sd{0.8} & \scorecell{50.8}{}\sd{3.3} & \scorecell{48.0}{}\sd{4.7} & \scorecell{48.6}{}\sd{1.0} & \scorecell{52.5}{}\sd{0.3} & \scorecell{45.4}{**}\sd{2.3} & \scorecell{49.6}{}\sd{1.0} & \scorecell{51.2}{}\sd{3.7} & \scorecell{50.3}{}\sd{0.4} & \scorecell{54.4}{**}\sd{3.3} \\
\midrule
\multirow{2}{*}{Binding} & principle\_a\_command & \scorecell{39.1}{***}\sd{5.9} & \scorecell{38.8}{***}\sd{5.3} & \scorecell{42.8}{***}\sd{2.1} & \scorecell{35.1}{***}\sd{4.5} & \scorecell{35.8}{}\sd{14.5} & \scorecell{40.9}{***}\sd{4.0} & \scorecell{36.7}{***}\sd{6.6} & \scorecell{44.1}{***}\sd{2.6} & \scorecell{47.4}{}\sd{5.4} & \scorecell{64.7}{***}\sd{6.2} & \scorecell{41.3}{***}\sd{1.7} \\
  & principle\_a\_locality & \scorecell{60.4}{***}\sd{1.2} & \scorecell{63.7}{***}\sd{1.7} & \scorecell{70.1}{***}\sd{0.7} & \scorecell{68.2}{***}\sd{1.6} & \scorecell{67.6}{***}\sd{4.8} & \scorecell{71.5}{***}\sd{1.6} & \scorecell{74.1}{***}\sd{2.9} & \scorecell{75.2}{***}\sd{1.8} & \scorecell{73.7}{***}\sd{4.9} & \scorecell{87.1}{***}\sd{2.8} & \scorecell{80.8}{***}\sd{1.1} \\
\midrule
\multirow{1}{*}{Wanna} & wanna & \scorecell{65.3}{}\sd{19.3} & \scorecell{74.3}{***}\sd{6.0} & \scorecell{7.6}{***}\sd{3.7} & \scorecell{78.1}{***}\sd{5.6} & \scorecell{82.1}{***}\sd{8.3} & \scorecell{38.9}{***}\sd{6.4} & \scorecell{63.6}{***}\sd{6.7} & \scorecell{71.7}{***}\sd{17.2} & \scorecell{52.0}{}\sd{20.4} & \scorecell{82.1}{***}\sd{3.0} & \scorecell{53.2}{}\sd{2.0} \\
\midrule
Average &   & \scorecell{61.0}{}\sd{1.6} & \scorecell{63.3}{}\sd{1.1} & \scorecell{54.3}{}\sd{1.5} & \scorecell{63.3}{}\sd{0.7} & \scorecell{62.5}{}\sd{1.9} & \scorecell{58.9}{}\sd{1.6} & \scorecell{62.3}{}\sd{2.6} & \scorecell{63.5}{}\sd{2.3} & \scorecell{61.3}{}\sd{1.9} & \scorecell{68.6}{}\sd{1.1} & \scorecell{64.3}{}\sd{2.1} \\
\bottomrule
\end{tabular}
\end{adjustbox}
\caption{Fine-grained PoSH results grouped by training size (columns) and phenomenon (rows) for LSTMs.}
\label{tab:finegrained-lstm}
\end{table*}

\begin{table*}[!th]
\centering
\begin{adjustbox}{max width=1\textwidth}
\begin{tabular}{ll*{11}{l}}
\toprule
  &   & \multicolumn{3}{c}{\textbf{10M}} & \multicolumn{3}{c}{\textbf{30M}} & \multicolumn{3}{c}{\textbf{50M}} & \multicolumn{2}{c}{\textbf{100M}} \\
\cmidrule(lr){3-5}\cmidrule(lr){6-8}\cmidrule(lr){9-11}\cmidrule(lr){12-13}
\textbf{Category} & \textbf{Phenomenon} & \babyfiltered & \baby & \textsc{wiki} & \babyfiltered & \baby & \textsc{wiki} & \babyfiltered & \baby & \textsc{wiki} & \baby & \textsc{wiki} \\
\midrule
\multirow{3}{*}{Island} & island\_adjunct & \scorecell{72.4}{***} & \scorecell{54.0}{} & \scorecell{70.8}{***} & \scorecell{82.8}{***} & \scorecell{59.2}{***} & \scorecell{69.4}{***} & \scorecell{82.0}{***} & \scorecell{61.0}{***} & \scorecell{70.2}{***} & \scorecell{80.4}{***} & \scorecell{51.6}{} \\
  & island\_complex\_np & \scorecell{25.6}{***} & \scorecell{34.8}{***} & \scorecell{48.0}{} & \scorecell{25.0}{***} & \scorecell{35.8}{***} & \scorecell{49.6}{} & \scorecell{27.2}{***} & \scorecell{55.0}{} & \scorecell{55.2}{} & \scorecell{36.2}{***} & \scorecell{48.2}{} \\
  & island\_wh & \scorecell{76.8}{***} & \scorecell{78.6}{***} & \scorecell{72.0}{***} & \scorecell{74.6}{***} & \scorecell{93.0}{***} & \scorecell{77.0}{***} & \scorecell{75.0}{***} & \scorecell{100.0}{***} & \scorecell{75.2}{***} & \scorecell{75.0}{***} & \scorecell{67.4}{***} \\
\midrule
\multirow{3}{*}{Question Formation} & question\_formation\_or & \scorecell{60.2}{***} & \scorecell{53.4}{} & \scorecell{59.6}{***} & \scorecell{59.0}{***} & \scorecell{53.8}{} & \scorecell{64.6}{***} & \scorecell{62.4}{***} & \scorecell{50.2}{} & \scorecell{64.4}{***} & \scorecell{62.2}{***} & \scorecell{46.2}{} \\
  & question\_formation\_rr & \scorecell{60.4}{***} & \scorecell{51.4}{} & \scorecell{60.8}{***} & \scorecell{61.6}{***} & \scorecell{51.8}{} & \scorecell{62.4}{***} & \scorecell{60.0}{***} & \scorecell{50.6}{} & \scorecell{64.0}{***} & \scorecell{63.0}{***} & \scorecell{49.0}{} \\
  & question\_formation\_sr & \scorecell{61.6}{***} & \scorecell{49.8}{} & \scorecell{52.2}{} & \scorecell{49.8}{} & \scorecell{52.6}{} & \scorecell{47.8}{} & \scorecell{48.4}{} & \scorecell{49.6}{} & \scorecell{44.6}{} & \scorecell{47.2}{} & \scorecell{45.8}{} \\
\midrule
\multirow{2}{*}{Binding} & principle\_a\_command & \scorecell{50.2}{} & \scorecell{50.0}{} & \scorecell{50.0}{} & \scorecell{50.2}{} & \scorecell{50.0}{} & \scorecell{49.6}{} & \scorecell{50.4}{} & \scorecell{49.6}{} & \scorecell{49.6}{} & \scorecell{49.8}{} & \scorecell{50.0}{} \\
  & principle\_a\_locality & \scorecell{40.4}{***} & \scorecell{40.8}{***} & \scorecell{37.6}{***} & \scorecell{36.0}{***} & \scorecell{42.6}{**} & \scorecell{38.0}{***} & \scorecell{35.8}{***} & \scorecell{42.8}{**} & \scorecell{38.8}{***} & \scorecell{37.0}{***} & \scorecell{47.8}{} \\
\midrule
\multirow{1}{*}{Wanna} & wanna & \scorecell{77.4}{***} & \scorecell{66.8}{***} & \scorecell{53.0}{} & \scorecell{79.6}{***} & \scorecell{58.6}{***} & \scorecell{49.8}{} & \scorecell{74.6}{***} & \scorecell{5.8}{***} & \scorecell{49.8}{} & \scorecell{73.6}{***} & \scorecell{56.2}{*} \\
\midrule
Average &   & \scorecell{58.3}{} & \scorecell{53.3}{} & \scorecell{56.0}{} & \scorecell{57.6}{} & \scorecell{55.3}{} & \scorecell{56.5}{} & \scorecell{57.3}{} & \scorecell{51.6}{} & \scorecell{56.9}{} & \scorecell{58.3}{} & \scorecell{51.4}{} \\
\bottomrule
\end{tabular}
\end{adjustbox}
\caption{Fine-grained PoSH results grouped by training size (columns) and phenomenon (rows) for 5-gram models.}
\label{tab:finegrained-ngram}
\end{table*}

\begin{table*}[]
\small
\centering
\begin{adjustbox}{width=0.99\textwidth}
\begin{tabular}{l|ccc|ccc|ccc|cc}
\toprule
Model & \multicolumn{3}{c|}{10M}
& \multicolumn{3}{c|}{30M}
& \multicolumn{3}{c|}{50M}
& \multicolumn{2}{c}{100M} \\
\cmidrule(lr){2-4}
\cmidrule(lr){5-7}
\cmidrule(lr){8-10}
\cmidrule(lr){11-12}
& \baby & \babyfiltered & \wiki
& \baby & \babyfiltered & \wiki
& \baby & \babyfiltered & \wiki
& \baby & \wiki \\
\midrule
Transformer & 139.0 & 132.4 &152.1 & 92.6& 92.7& 60.8 &78.4 &78.8 & 75.0& 30.0& 47.1 \\
LSTM        & 160.5&153.5 &178.8 &110.9 & 125.3&97.1 & 95.2 &97.1 & 88.4& 42.0& 61.2\\
5-gram      & 251.6 &251.7 &338.0 & 199.0&199.2 & 182.1&185.5 &185.6 &212.1 & 89.7&150.2 \\
\bottomrule
\end{tabular}
\end{adjustbox}
\caption{Perplexity on the validation split of each dataset for different models. Texts within the same data split are tokenized using the same tokenizer. Perplexity values are comparable only within a split and should not be compared across different data splits due to different tokenizers used.}
\label{tab:ppl}
\end{table*}

In Experiment~3, the dynamic recency bias condition was trained and evaluated by \textit{epochs} rather than by gradient steps, due to the implementation of the time-dependent bias update.  
We adopted our standard GPT-2 mini architecture but followed the hyperparameter configuration of \citet{mita-etal-2025-developmentally} for comparability.  
Specifically, the maximum context length was set to 32 and the batch size to 512, as these settings yielded stable learning and positive performance in their reported experiments.

\begin{table*}[tbp]
\small
    \centering
    \begin{tabular}{l|lllllll}
    \toprule
    Vocab size  & Dataset  &  10M & 30M & 50M & 100M  \\
    \midrule
    \multirow{2}{*}{8192}& \textsc{wiki} & 17,341,898 & 54,230,974 & 90,741,059 & 181,490,178\\
  & baby & 16,216,427&  50,045,572 & 83,924,585 & 159,613,479\\
  \midrule 
   \multirow{2}{*}{32768}& \textsc{wiki} & 14,442,880 & 45,544,347 & 76,384,301 & 153,006,220\\
  & baby &  153,18,669& 46,823,177 &   78,091,016 & 146,768,608\\
  \midrule 
   \multirow{2}{*}{49152}& \textsc{wiki} & 13,989,838 & 44,114,584 & 74,020,389 & 148,359,082\\
  & baby &  15,181,258 & 46,370,173 & 77,257,389 & 144,939,275\\
  \midrule 
    \multirow{2}{*}{65536}& \textsc{wiki} & 13,741,202 & 43,308,251 & 72,687,684 & 145,743,672\\
  & baby & 15,114,124 & 46,128,756 & 76,809,625& 143,962,866\\
  \bottomrule
    \end{tabular}
    \caption{The corpus token count (CTC) of BPE tokenizers with different vocab size trained with different datasets}
    \label{tab:ctc}
\end{table*}

\begin{table*}[tbp]
\small
\centering
\begin{tabular}{llp{8cm}}
\toprule
Category & Benchmark & Phenomena \\
\midrule
\multirow{5}{*}{Island}
& BLiMP
& \texttt{adjunct\_island}, \texttt{wh\_island}, \texttt{complex\_NP\_island} \\
& SCaMP (plausible)
& \texttt{complex\_np\_island}, \texttt{wh\_island}, \texttt{adjunct\_island} \\
& SCaMP (implausible)
& \texttt{complex\_np\_island\_implausible}, \texttt{wh\_island\_implausible}, \texttt{adjunct\_island\_implausible} \\
& Zorro
& \texttt{island-effects-adjunct\_island} \\
& PoSH
& \texttt{island-adjunct}, \texttt{island-complex-np}, \texttt{island-wh} \\
\midrule
\multirow{1}{*}{Question Formation}
& PoSH
& \texttt{question-formation\_or}, \texttt{question-formation\_rr}, \texttt{question-formation\_sr} \\
\midrule
\multirow{1}{*}{Wanna}
& PoSH
& \texttt{wanna} \\
\midrule
\multirow{5}{*}{Binding}
& BLiMP
& \texttt{principle\_A\_c\_command}, \texttt{principle\_A\_case\_2},\texttt{principle\_A\_domain\_2}, \texttt{principle\_A\_domain\_3} \\
& Zorro
& \texttt{binding-principle\_a} \\
& SCaMP (plausible)
& \texttt{principle\_A\_domain\_2}, \texttt{principle\_A\_domain\_3}, \texttt{principle\_A\_c\_command} \\
& SCaMP (implausible)
& \texttt{principle\_A\_domain\_2\_implausible}, \texttt{principle\_A\_domain\_3\_implausible}, \texttt{principle\_A\_c\_command\_implausible} \\
& PoSH
& \texttt{principle\_a\_command}, \texttt{principle\_a\_locality} \\
\bottomrule
\end{tabular}
\caption{Mapping between syntactic categories, benchmarks, and corresponding evaluation phenomena.}
\label{tab:category-dataset-map}
\end{table*}

\begin{figure*}
    \centering\includegraphics[width=\linewidth]{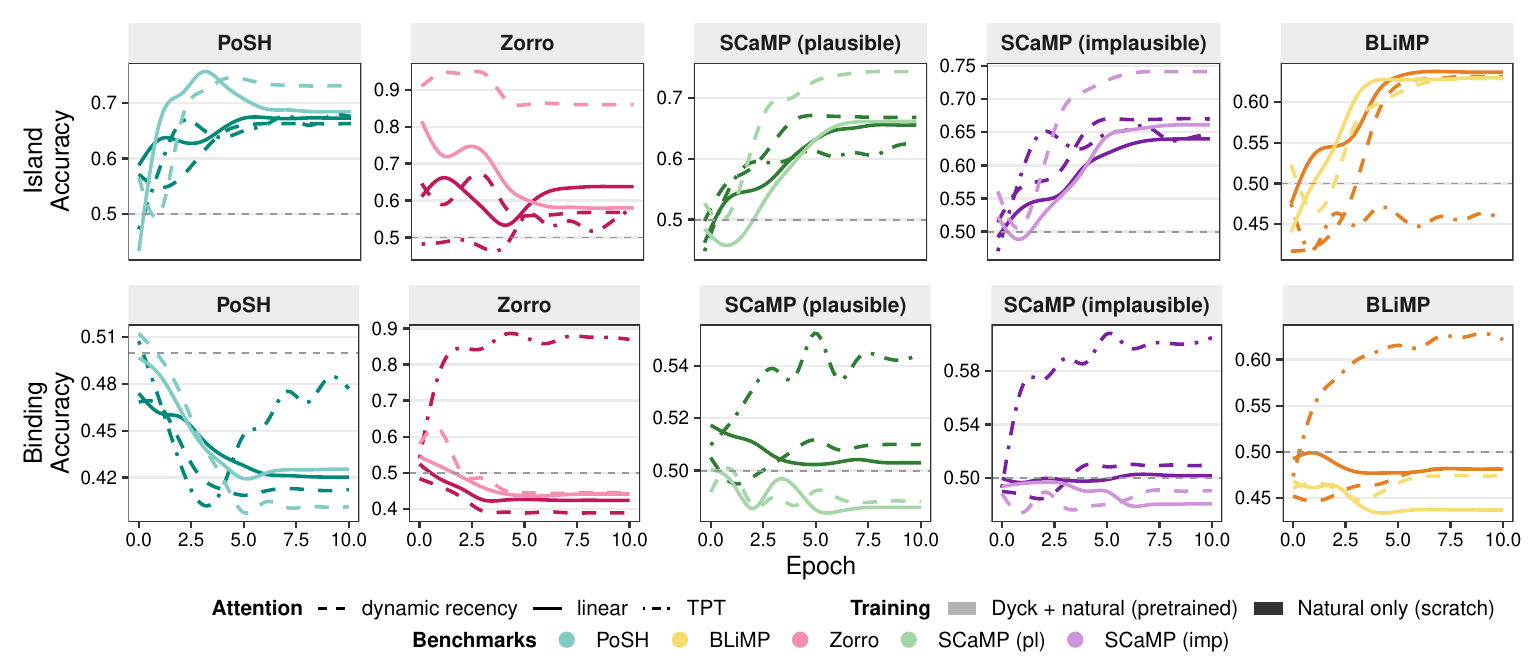}
    \caption{Performance of different models on selected PoS islands and binding phenomena in each benchmark (the performance is average of three random seeds)}
    \label{fig:islandandbinding}
\end{figure*}


\section{Benchmark Construction Pipeline}
\label{benchmark-pip}
We follow \citet{huebner-etal-2021-babyberta} and \citet{warstadt-etal-2020-blimp-benchmark} and use templates to generate minimal pairs. Each subcategory has at least two templates as shown in \Cref{tab:full-templates}.

\begin{table*}[!th]
\centering
\begin{subtable}{0.48\textwidth}

\begin{adjustbox}{width=\textwidth}
\begin{tabular}[t]{lcc}
\toprule
Predictor & Estimate & 95\% CrI\\
\midrule
Intercept & 0.63 & {}[0.06, 1.15]\\
Training size (z) & 0.16 & {}[0.14, 0.17]\\
Filtering (filtered vs. unfiltered) & 0.04 & {}[0.00, 0.08]\\
Training source (baby vs. wiki) & 0.27 & {}[0.24, 0.30]\\
Category: Binding & -0.11 & {}[-0.78, 0.53]\\
Category: Islands & 0.32 & {}[-0.35, 0.93]\\
Category: QF & -0.25 & {}[-0.86, 0.42]\\
\addlinespace
Category: Wanna & 0.06 & {}[-0.91, 1.04]\\
\midrule
\textit{Interaction Effects} &  & \\
Training size (z) $\times$ Filtering & -0.02 & {}[-0.07, 0.02]\\
\addlinespace
Training size (z) $\times$ Domain & 0.04 & {}[0.01, 0.06]\\
Binding $\times$ Filtering & -0.05 & {}[-0.11, 0.01]\\
Islands $\times$ Filtering & -0.23 & {}[-0.29, -0.17]\\
QF $\times$ Filtering & 0.14 & {}[0.09, 0.18]\\
Wanna $\times$ Filtering & 0.31 & {}[0.22, 0.40]\\
\addlinespace
Binding $\times$ Domain & 0.17 & {}[0.11, 0.22]\\
Islands $\times$ Domain & 0.11 & {}[0.06, 0.16]\\
QF $\times$ Domain & -0.26 & {}[-0.31, -0.22]\\
Wanna $\times$ Domain & 1.07 & {}[0.99, 1.14]\\
\midrule
\textit{Group-level standard deviations} &  & \\
\addlinespace
Phenomenon intercept SD & 0.76 & {}[0.46, 1.27]\\
Random-seed intercept SD & 0.03 & {}[0.00, 0.17]\\
\bottomrule
\end{tabular}
\end{adjustbox}
\caption{Transformer}
\label{tab:transformer}
\end{subtable}
\hfill
\begin{subtable}{0.48\textwidth}
\centering
\begin{adjustbox}{width=\textwidth}
\begin{tabular}[t]{lcc}
\toprule
Predictor & Estimate & 95\% CrI\\
\midrule
Intercept & 0.54 & {}[-0.02, 1.04]\\
Training size (z) & 0.10 & {}[0.08, 0.11]\\
Filtering (filtered vs. unfiltered) & -0.15 & {}[-0.18, -0.11]\\
Training source (baby vs. wiki) & 0.49 & {}[0.46, 0.52]\\
Category: Binding & -0.12 & {}[-0.74, 0.54]\\
Category: Islands & 0.48 & {}[-0.21, 1.04]\\
Category: QF & -0.32 & {}[-0.90, 0.36]\\
\addlinespace
Category: Wanna & -0.04 & {}[-0.96, 0.88]\\

\midrule
\textit{Interaction Effects} &  & \\
Training size (z) $\times$ Filtering & -0.05 & {}[-0.10, -0.01]\\
\addlinespace
Training size (z) $\times$ Domain & -0.05 & {}[-0.07, -0.02]\\
Binding $\times$ Filtering & -0.29 & {}[-0.35, -0.23]\\
Islands $\times$ Filtering & 0.05 & {}[-0.01, 0.11]\\
QF $\times$ Filtering & 0.06 & {}[0.01, 0.11]\\
Wanna $\times$ Filtering & -0.41 & {}[-0.49, -0.32]\\
\addlinespace
Binding $\times$ Domain & 0.06 & {}[0.00, 0.11]\\
Islands $\times$ Domain & 0.37 & {}[0.32, 0.42]\\
QF $\times$ Domain & -0.22 & {}[-0.26, -0.18]\\
Wanna $\times$ Domain & 1.74 & {}[1.67, 1.82]\\
\midrule
\textit{Group-level standard deviations} &  & \\
\addlinespace
Phenomenon intercept SD & 0.71 & {}[0.42, 1.23]\\
Random-seed intercept SD & 0.03 & {}[0.00, 0.15]\\
\bottomrule
\end{tabular}
\end{adjustbox}
\caption{LSTM}
\label{tab:lstm}
\end{subtable}
\caption{Bayesian mixed-effects logistic regression predicting item-level accuracy.
Entries are posterior medians with 95\% credible intervals on the log-odds scale. Training size was $z$-scored. Category predictors are deviation-coded using sum-to-zero contrasts; thus, the coefficient for \textit{Wanna} is implicit and is recovered as the negative sum of the remaining category coefficients.
Positive estimates for two-level factors indicate higher accuracy for the first level (filtered vs.\ unfiltered; baby vs.\ wiki).}
\label{tab:brmsresults}
\end{table*}

\section{Benchmark design considerations}
\label{sec:benchmark-design}
\paragraph{Yes-no question} In \poshbench, we standardize evaluation using minimal pairs across three subordinate clause types tested in previous acquisition studies \citep{crain1987structure}: subject, object, and reduced relative clauses. 
\paragraph{Islands} In \poshbench, we include three subtypes, Complex NP, Wh-, and Adjunct Islands, all drawn from the acquisition literature.
\paragraph{Binding} In \poshbench, we therefore include two subtypes, locality and c-command, to assess sensitivity to structural binding constraints
\paragraph{\textit{Wanan}} In our minimal-pair design, we do not contrast grammatical \textit{want to} forms with ungrammatical \textit{wanna} forms (e.g., \textit{Who do you want to/*wanna go?}), because \textit{want to} is substantially more frequent than \textit{wanna} in all training corpora, introducing a frequency confound unrelated to structural generalization. Instead, we test whether models learn the structural restriction that \textit{wanna} may only precede intransitive verbs or transitive verb phrases (see Example~\ref{example-complex-wanna}).

\section{Bayesian mixed model results}
The Baysian mixed model results are reported in \Cref{tab:brmsresults}.

\section{\texttt{GPT5.4} Results}
We evaluate \texttt{GPT5.4} on PoSH-Bench to estimate the upper-bound performance of LLMs. We use zero-shot prompting with the following prompt: \textit{Which sentence is grammatical? A. {Sentence 1} B. {Sentence 2} Respond with exactly one character: A or B.} We randomize the order of the two sentences to mitigate potential bias toward option A. The results are reported in Table~\ref{tab:gpt5}.

\section{Overall \& Fine-grained Results}
\label{sec:finegrained}
\begin{figure*}[!th]
    \centering
    \includegraphics[width=0.9\linewidth]{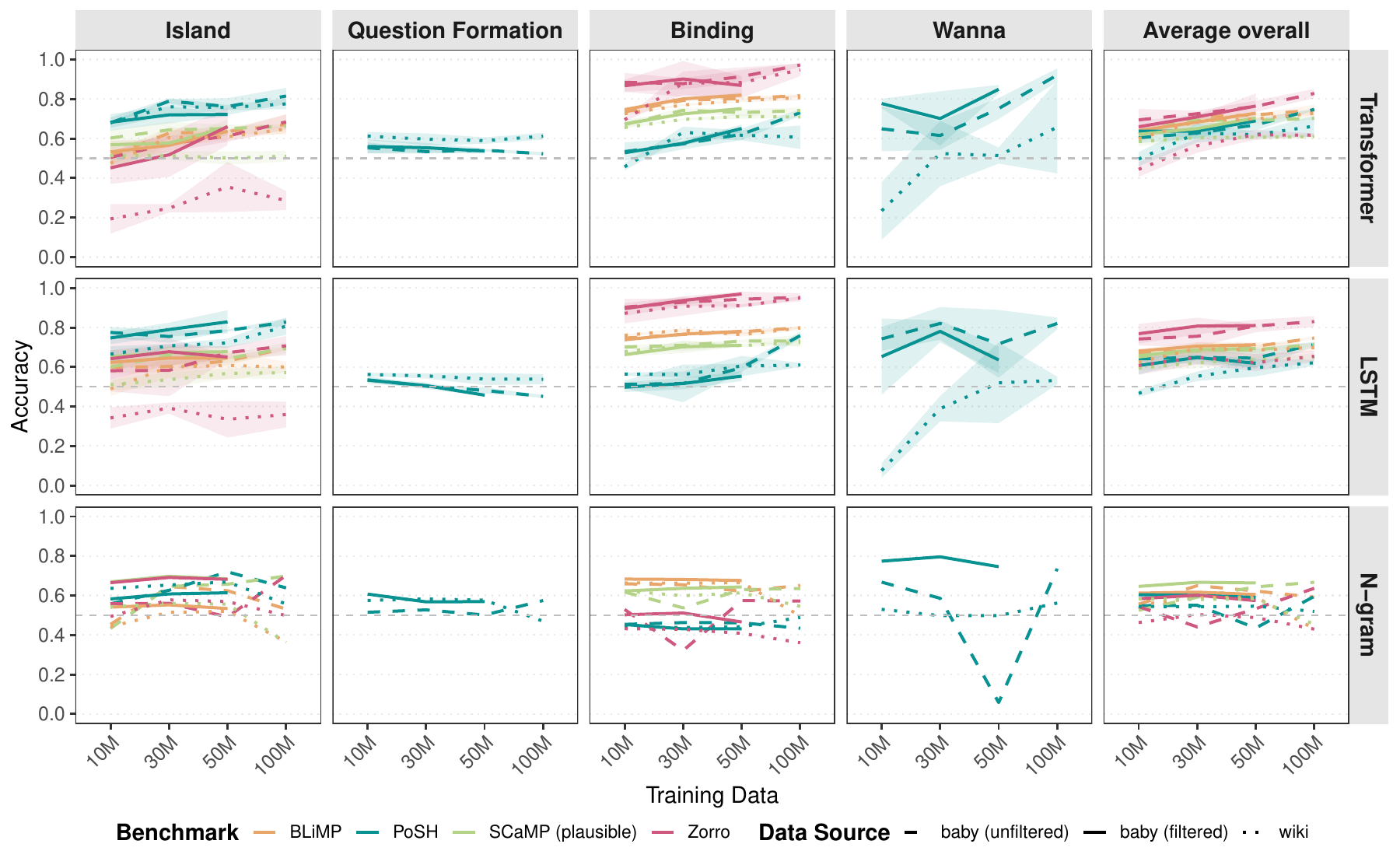}
    \caption{Poverty of the Stimulus Category-wise average for each benchmark. The horizontal dashed line represents chance-level performance. The shadow for transformer and LSTM models represents SD across 3 random seeds.}
\label{fig:overall}
\end{figure*}

The overall results across models and benchmarks are reported in \Cref{fig:overall} and subcategory-wise results can be found in \Cref{tab:finegrained,tab:finegrained-lstm,tab:finegrained-ngram}.

\begin{table*}[!th]
\centering
\scriptsize
\setlength{\tabcolsep}{6pt}
\begin{adjustbox}{max width=0.9\textwidth}
\begin{tabular}{p{0.22\textwidth} p{0.78\textwidth}}
\toprule
Category & Template (bad/ungrammatical vs.\ good/grammatical) \\
\midrule
\multirow{4}{*}{Binding (c-command)}
& \texttt{b: \{nn\_m\} \{vb\_state\} that \{name\_m2\}'s \{nn\_f1\} \{vb\_trans\_animate\} himself.} \\
& \texttt{g: \{nn\_m\} \{vb\_state\} that \{name\_m2\}'s \{nn\_f1\} \{vb\_trans\_animate\} herself.} \\[0.3em]
\cmidrule(lr){2-2}
& \texttt{b: \{nn\_f\} \{vb\_state\} that \{name\_f2\}'s \{nn\_m1\} \{vb\_trans\_animate\} herself.} \\
& \texttt{g: \{nn\_f\} \{vb\_state\} that \{name\_f2\}'s \{nn\_m1\} \{vb\_trans\_animate\} himself.} \\[0.3em]
\cmidrule(lr){2-2}
& \texttt{b: \{nn\_f\} \{vb\_state\} that the \{nn\_m2\} who \{vb\_trans\_animate\} \{det\}\{nn\_f\} \{vb2\} herself.} \\
& \texttt{g: \{nn\_f\} \{vb\_state\} that the \{nn\_m2\} who \{vb\_trans\_animate\} \{det\}\{nn\_f\} \{vb2\} himself.} \\[0.3em]
\cmidrule(lr){2-2}
& \texttt{b: \{nn\_m\} \{vb\_state\} that the \{nn\_f2\} who \{vb\_trans\_animate\} \{det\}\{nn\_m\} \{vb2\} himself.} \\
& \texttt{g: \{nn\_m\} \{vb\_state\} that the \{nn\_f2\} who \{vb\_trans\_animate\} \{det\}\{nn\_m\} \{vb2\} herself.} \\

\midrule

\multirow{4}{*}{Binding (locality)}
& \texttt{b: \{nn\_f\} \{vb\_state\} that the \{nn\_m1\} \{vb\_trans\_animate\} herself.} \\
& \texttt{g: \{nn\_f\} \{vb\_state\} that the \{nn\_m1\} \{vb\_trans\_animate\} himself.} \\[0.3em]
\cmidrule(lr){2-2}
& \texttt{b: \{nn\_m\} \{vb\_state\} that the \{nn\_f1\} \{vb\_trans\_animate\} himself.} \\
& \texttt{g: \{nn\_m\} \{vb\_state\} that the \{nn\_f1\} \{vb\_trans\_animate\} herself.} \\[0.3em]
\cmidrule(lr){2-2}
& \texttt{b: \{nn\_f\} \{vb\_percep\} herself \{vb\_ed\} \{det\} \{nn\}.} \\
& \texttt{g: \{nn\_f\} \{vb\_percep\} herself \{vb\_gerund\} \{det\} \{nn\}.} \\[0.3em]
\cmidrule(lr){2-2}
& \texttt{b: \{nn\_m\} \{vb\_percep\} himself \{vb\_ed\} \{det\} \{nn\}.} \\
& \texttt{g: \{nn\_m\} \{vb\_percep\} himself \{vb\_gerund\} \{det\} \{nn\}.} \\

\midrule

\multirow{4}{*}{Adjunct island}
& \texttt{b: Who should \{det\} \{person\} \{trans\_verb1\} \{nn\_person\} \{pp\} \{trans\_verb\_gerund\}?} \\
& \texttt{g: Who should \{det\} \{person\} \{trans\_verb1\} \{pp\} \{trans\_verb\_gerund\} \{nn\_person\}?} \\[0.3em]
\cmidrule(lr){2-2}
& \texttt{b: What did \{det\} \{person\} \{trans\_verb2\} the \{nn\_obj\} \{pp\} \{trans\_verb\_gerund\}?} \\
& \texttt{g: What did \{det\} \{person\} \{trans\_verb2\} \{pp\} \{trans\_verb\_gerund\} the \{nn\_obj\}?} \\

\midrule

\multirow{4}{*}{Complex NP island}
& \texttt{b: who did \{det\}  \{person\} who \{trans\_verb\} \{trans\_verb2\} \{person2\}?} \\
& \texttt{g: who did \{det\} \{person\} who \{trans\_verb\} \{person2\} \{trans\_verb2\}?} \\[0.3em]
\cmidrule(lr){2-2}
& \texttt{b: what did \{det\} \{person\} who \{trans\_verb\} \{trans\_verb2\} \{person2\}?} \\
& \texttt{g: what did \{det\} \{person\} who \{trans\_verb\} \{person2\} \{trans\_verb2\}?} \\

\midrule

\multirow{8}{*}{Wh-island}
& \texttt{b: who did \{det\} \{person\} \{vb\_state\} whether \{name\} \{vb\_trans1\}?} \\
& \texttt{g: who did \{det\} \{person\} \{vb\_state\} that \{name\} \{vb\_trans1\}?} \\[0.3em]
\cmidrule(lr){2-2}
& \texttt{b: what did \{det\} \{person\} \{vb\_state\} whether \{name\} \{vb\_trans\}?} \\
& \texttt{g: what did \{det\} \{person\} \{vb\_state\} that \{name\} \{vb\_trans\}?} \\[0.3em]
\cmidrule(lr){2-2}
& \texttt{b: what \{aux\} \{det\} \{person\} \{vb\_question\} who \{vb\_trans\}?} \\
& \texttt{g: what \{aux\} \{det\} \{person\} \{vb\_question\} \{pron\} \{vb\_trans\}?} \\[0.3em]
\cmidrule(lr){2-2}
& \texttt{b: what \{aux\} \{det\} \{person\} \{vb\_question\} who \{vb\_trans2\}?} \\
& \texttt{g: what \{aux\} \{det\} \{person\} \{vb\_question\} \{pron\} \{vb\_trans2\}?} \\

\midrule

\multirow{4}{*}{Question formation (Obj-Rel)}
& \texttt{b: \{aux2\} \{det\} \{nn\_human\} who \{det2\} \{noun1\} \{verb1\} \{aux1\} \{verb2\} \{det3\} \{noun2\}?} \\
& \texttt{g: \{aux1\} \{det\} \{nn\_human\} who \{det2\} \{noun1\} \{aux2\} \{verb1\} \{verb2\} \{det3\} \{noun2\}?} \\[0.3em]
\cmidrule(lr){2-2}
& \texttt{b: \{aux2\} \{det\} \{nn\_human\} who \{det2\} \{noun1\} \{verb1\} \{aux1\} \{verb\_intrans\}?} \\
& \texttt{g: \{aux1\} \{det\} \{nn\_human\} who \{det2\} \{noun1\} \{aux2\} \{verb1\} \{verb\_intrans\}?} \\

\midrule

\multirow{4}{*}{Question formation (Subj-Rel)}
& \texttt{b: \{aux2\} \{det\} \{nn\_human\} who \{verb1\} the \{noun1\} \{aux1\} \{verb2\} \{det2\} \{noun2\}?} \\
& \texttt{g: \{aux1\} \{det\} \{nn\_human\} who \{aux2\} \{verb1\} the \{noun1\} \{verb2\} \{det2\} \{noun2\}?} \\[0.3em]
\cmidrule(lr){2-2}
& \texttt{b: \{aux2\} \{det\} \{nn\_human\} who \{verb1\} \{det2\} the \{noun1\} \{aux1\} \{verb\_intrans\}?} \\
& \texttt{g: \{aux1\} \{det\} \{nn\_human\} who \{aux2\} \{verb1\} \{det2\} the \{noun1\} \{verb\_intrans\}?} \\

\midrule
\multirow{4}{*}{Question formation (Red. Rel)}
& \texttt{b: \{aux2\} \{det\} \{nn\_human\} \{det2\} \{noun1\} \{verb1\} \{aux1\} \{verb2\} the \{noun2\}?} \\
& \texttt{g: \{aux1\} \{det\} \{nn\_human\} \{det2\} \{noun1\} \{aux2\} \{verb1\} \{verb2\} the \{noun2\}?} \\[0.3em]
\cmidrule(lr){2-2}
& \texttt{b: \{aux2\} \{det\} \{nn\_human\} \{det2\} \{noun1\} \{verb1\} \{aux1\} \{verb\_intrans\}?} \\
& \texttt{g: \{aux1\} \{det\} \{nn\_human\} \{det2\} \{noun1\} \{aux2\} \{verb1\} \{verb\_intrans\}?} \\

\midrule

\multirow{2}{*}{Wanna}
& \texttt{b: \{who|what\} \{aux\_sg\} the \{adj\_person\} \{nn1\_human\} wanna \{vb\_trans\_animate\} the \{nn2\_obj\}?} \\
& \texttt{g: \{when|how|why|where\} \{aux\_sg\} the \{adj\_person\} \{nn1\_human\} wanna \{vb\_trans\_animate\} the \{nn2\_obj\}?} \\[0.3em]
\cmidrule(lr){2-2}
& \texttt{b: \{who|what\} \{aux\_sg\} \{name\} wanna \{vb\_intrans\_animate\}?} \\
& \texttt{g: \{when|how|why|where\} \{aux\_sg\} \{name\} wanna \{vb\_intrans\_animate\}?} \\
\bottomrule
\end{tabular}
\end{adjustbox}
\caption{Complete minimal-pair templates used to generate all items. \texttt{b} = bad, \texttt{g} = good.}
\label{tab:full-templates}
\end{table*}
\end{document}